\documentclass[10pt,conference]{IEEEtran}
\IEEEoverridecommandlockouts
\usepackage{cite}
\usepackage{amsmath,amssymb,amsfonts}
\usepackage{graphicx}
\usepackage{textcomp}
\usepackage{xcolor}

\usepackage{microtype}
\usepackage{subfigure}
\usepackage{booktabs} % for professional tables

\usepackage{tikz}
\usepackage{multirow}
\usepackage{enumitem}
\usepackage{tabularray}
\usepackage{listings}
\usepackage{hyperref}
\usepackage{algorithm}
\usepackage{tabularray}

\usepackage{algpseudocode}

\newcommand{\name}{{\tt PP-DNN}}
\newcommand{\nameS}{{\tt PP-DNN }}

\def\BibTeX{{\rm B\kern-.05em{\sc i\kern-.025em b}\kern-.08em
    T\kern-.1667em\lower.7ex\hbox{E}\kern-.125emX}}
\begin{document}

\title{Enhancing Predictability of Multi-Tenant DNN Inference for Autonomous
Vehicles' Perception}

\author{
\IEEEauthorblockN{
Liangkai~Liu\IEEEauthorrefmark{1},
Kang~G.~Shin\IEEEauthorrefmark{1},
Jinkyu~Lee\IEEEauthorrefmark{2},
Chengmo~Yang\IEEEauthorrefmark{3}, and
Weisong~Shi\IEEEauthorrefmark{3}
}
\IEEEauthorblockA{
\IEEEauthorrefmark{1}Department of Electrical Engineering and Computer Science, University of Michigan
\\
\IEEEauthorrefmark{2}Department of Computer Science and Engineering, Yonsei University
\\
\IEEEauthorrefmark{3}Department of Computer and Information Sciences, University of Delaware\\
}
}

\maketitle

\begin{abstract}
Autonomous vehicles (AVs) rely on sensors and deep neural networks 
(DNNs) to perceive their surrounding environment and make 
maneuver decisions in real time. However, achieving 
real-time DNN inference in the AV's perception pipeline is 
challenging due to the large gap between the computation 
requirement and the AV's limited resources. 
Most, if not all, of existing studies focus on optimizing the 
DNN inference time to achieve faster perception, for example, 
by compressing the DNN model with pruning and quantization.

In contrast, we present a Predictable Perception system with 
DNNs (\name) that reduce the amount of image data to be processed
while maintaining the same level of accuracy for multi-tenant 
DNNs by dynamically selecting `critical' frames and regions 
of interest (ROIs).
\name\ is based on our key insight that critical frames and 
ROIs for AVs vary with the AV's surrounding environment.
However, it is challenging to identify and use critical frames 
and ROIs in multi-tenant DNNs for predictable inference. 
Given image-frame streams, \name\ leverages an ROI generator 
to identify critical frames and ROIs based on the similarities 
of consecutive frames and traffic scenarios. 
\name\ then leverages a FLOPs predictor to predict 
multiply-accumulate operations (MACs) from the dynamic 
critical frames and ROIs. The ROI scheduler coordinates 
the processing of critical frames and ROIs with multiple 
DNN models. Finally, we design a detection predictor
for the perception of non-critical frames.
We have implemented \name\ in an ROS-based AV pipeline 
and evaluated it with the BDD100K and the nuScenes dataset. 
\name\ is observed to significantly enhance perception 
predictability, increasing the number of fusion frames 
by up to 7.3$\times$, reducing the fusion delay by 
$>$2.6$\times$ and fusion-delay variations by $>$2.3$\times$, 
improving detection completeness by 75.4\% and the 
cost-effectiveness by up to 98\% over the baseline.
\end{abstract}

%\begin{IEEEkeywords}
%autonomous vehicles, predictability, deep neural networks
%\end{IEEEkeywords}

\section{Introduction}

% Why the target problem is important?

Deep neural networks (DNNs) have been widely used in the perception 
pipeline of autonomous vehicles (AVs) thanks to their high accuracy 
and ability to learn from raw data~\cite{yurtsever2020survey,kato2015open}. For example, YOLOv3, 
SSD, and Faster R-CNN have been used for object detection~\cite{redmon2018yolov3,liu2016ssd,ren2015faster}. 
Deeplabv3+ was proposed for semantic segmentation~\cite{howard2017mobilenets, chen2018encoder}, while LaneNet 
was designed for lane detection~\cite{neven2018towards}. 
An AV's perception should be completed before a certain 
deadline with high accuracy~\cite{lin2018architectural,gog2022d3,liu2020computing}.

Existing efforts focus on model compression to speed up the 
execution of DNN inference. 
% One of them is to compress the DNN models.  
Han \textit{et al.}~\cite{han2015learning} proposed to prune 
redundant connections and re-train the deep learning models 
in order to fine-tune the weights effectively, thus reducing 
the computation requirement. Lowering the precision of 
operations and operands is another way of reducing the 
runtime of DNN inference~\cite{cai2017deep}. 
However, it usually accompanies non-negligible loss of 
accuracy, which is often unacceptable for AVs. 
Researchers also attempted to reduce the DNN inference time. 
DeepCache~\cite{xu2018deepcache} leverages the temporal locality in 
streaming video to accelerate the execution of vision tasks on 
mobile devices. ALERT~\cite{wan2020alert} used an anytime DNN system, 
which yields different outputs with different execution times. 
Liu \textit{et al.}~\cite{liu2022self} proposed a 
self-cueing attention mechanism in critical regions for 
real-time object tracking.
However, unlike general machine perception systems, the AV 
pipeline is usually equipped with multi-tenant DNNs 
running simultaneously, and hence their results need 
to be combined via fusion. 
Prophet~\cite{liuprophet} achieves timing predictability for 
multi-tenant DNNs \textit{without} any accuracy guarantee. 
Guaranteeing the predictability of 
the AV perception is still a difficult and open problem. 

Unlike the state-of-the-art (SOTA) studies that primarily focus 
on optimizing the DNN inference time to enhance accuracy, 
we seek a new and complementary opportunity: \textit{Can we 
reduce the number of image frames to be processed without loss 
of accuracy by dynamically adjusting critical frames and 
regions of interest (ROIs) ?} 
In AV perception, an image frame is said to be {\em critical} if 
it carries vital information that affects the AV's safety. 
A critical frame usually exhibits a noticeable difference from 
its preceding frames. We observe that the critical frames and 
ROIs for AVs vary with their surrounding environment. 
Given the temporal correlation within the continuous stream of images 
used for the detection of objects, processing every individual 
frame is not always necessary. 
The frequency of critical frames should instead be dynamically 
adapted to the specific driving and traffic conditions. 
Moreover, the configuration of ROI is also subject to 
environmental variations. For instance, when an 
AV stops at a traffic light, the ROI should cover all vehicles, 
pedestrians, bicycles, and other objects within its vicinity. 
On the other hand, the ROIs on a highway only need to cover the 
frontal and nearby vehicles.

We propose \name, an adaptive AV perception system, which dynamically 
selects critical frames and ROIs to guarantee predictability for 
multi-tenant DNN inference of objects in the AV's vicinity. 
For given image streams, \name\ first identifies critical frames 
and ROIs based on the similarity of consecutive frames in each 
image stream, box tracking, and traffic scenarios. 
It then coordinates the processing of critical frames and ROIs with 
multiple DNN models for environmental perception. As a result, \name\ 
not only reduces the computation requirement but also achieves lower 
inference and fusion latencies without loss of perception accuracy.

Unfortunately, it is difficult to identify critical frames 
and ROIs for AVs while they are moving. For their adaptive and 
robust selection, \name\ proposes an ROI generator to choose 
representative frames with multi-level criteria. It first analyzes 
the pixel-level differences of consecutive frames using the 
{\em structural similarity} (SSIM) index. 
A lightweight tracker is designed to track stationary and moving 
objects to determine the ROIs together with the driving conditions 
like speed and highway or downtown. 
By combining the critical frame candidates with ROIs, the ROI 
generator dynamically produces and delivers critical frames to 
the inference pipeline equipped with multi-tenant DNNs. 

It is difficult to process the selected critical frames with 
dynamic ROIs and multi-tenant DNNs because the variable ROI sizes 
cause the inference time to vary with the fusion of results from 
multiple DNN inference tasks, which may result in a longer delay 
or even failure of fusion. \name\ thus introduces a FLOPs predictor 
to predict Multiply-Accumulate Operations (MACs) for different ROIs. 
To reduce the fusion delay between multi-tenant DNNs, the frame 
scheduler monitors the delay for each task and determines 
the ROI for each task. Depending on the progress of fusion, 
it requests some tasks to yield resources or skip frames. 
Finally, \name\ introduces a detection predictor to generate 
anticipated results for non-critical frames, which decouples 
the detection from the fusion. The detection predictor 
takes the tracking results with selected ROIs to generate 
the velocity of bounding boxes for all moving objects. 
By combining it with the original image streams, \name\  
predicts the coordinates of objects, lanes, and segmentation. 

We have implemented \name\ in a Robot Operating System (ROS) 
based pipeline and evaluated it with real AV datasets: BDD100K 
and nuScenes \cite{yu2020bdd100k,caesar2020nuscenes}. Overall, this paper makes the following four main contributions:
\begin{itemize}
\item Discovery of two unique characteristics 
(\S\ref{sec:insights}) for building a predictable perception 
pipeline via empirical studies.

\item Design of the ROIs Generator (\S\ref{subsec:roi-generator}) 
to model and detect ROIs in dynamic environments and development 
of the Detection Predictor (\S\ref{subsec:detect-predict}) to 
predict detection results for non-critical frames using 
prior detections and temporal locality.

\item Design of a Task Coordinator (\S\ref{subsec:task-coordinate}) 
to dynamically select ROIs, optimize input sizes for all 
perception tasks, and coordinate the execution order of 
critical frames and ROIs in multi-tenant DNNs.

\item Addressing the lack of end-to-end testing by 
implementing \name{} atop ROS (\S\ref{sec:implementation}) and 
evaluating it with real-world AV datasets (\S\ref{sec:evaluation}), 
demonstrating that \name{} improves the number of fusion frames by 
up to 7.3$\times$, reduces fusion delay by $>2.6$$\times$ and 
fusion-delay variations by $>2.3$$\times$, improves
detection completeness by 75.4\% and cost-effectiveness 
by 98\% over the baseline with general settings.
\end{itemize}

% \section{DNNs in AV Pipeline}
\section{Background and Motivation}
\label{sec:motivation}

We first introduce DNNs in a general AV 
pipeline \cite{yurtsever2020survey,liu2020computing} and then
motivate the need for predictable perception . 
% Next, we highlight the opportunities that motivate our work.

\subsection{DNNs in AV Pipeline}
% \label{subsec:DNNs-AV} 
 
\begin{figure}[!htp]
	\centering
	\includegraphics[width=\columnwidth]{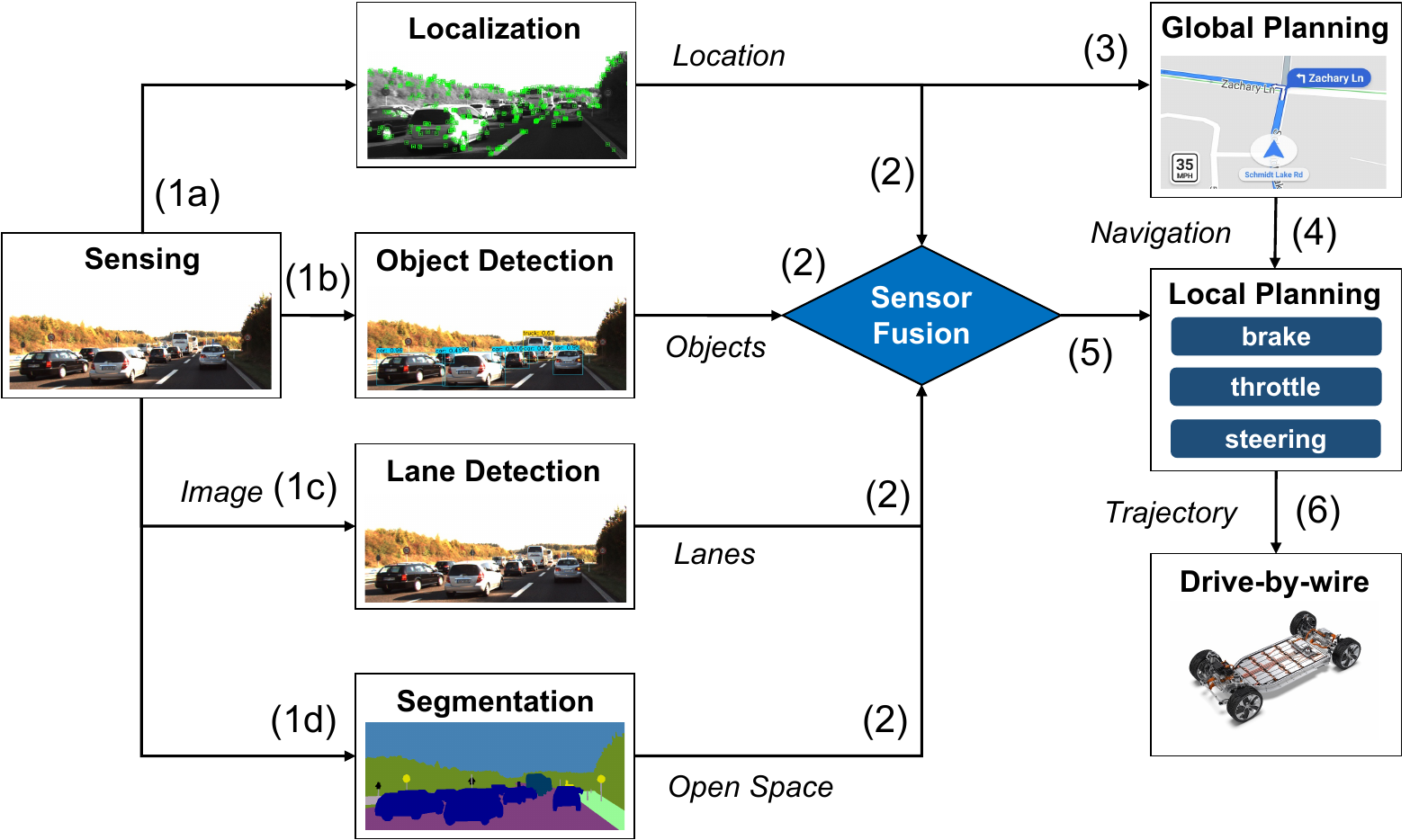}
        \vspace{-1em}
	\caption{A general autonomous driving pipeline.}
	\label{fig:e2e-pipeline}
\end{figure}

Fig.~\ref{fig:e2e-pipeline} shows a generalized pipeline for 
modular AVs. A sensing node publishes the captured sensor data 
to all the perception nodes for localization (step 1a)
\cite{mur2017orb}, object detection (step 1b)
\cite{ren2015faster,he2017mask,redmon2018yolov3}, lane detection 
(step 1c)~\cite{neven2018towards, ko2020key, zheng2020resa, 
pan2018spatial}, and segmentation (1d)~\cite{ronneberger2015u, 
chen2018encoder}. Next, the perception results are submitted to a 
sensor fusion node (step 2), which combines the information on 
the vehicle's location, objects, lanes, and open spaces
\cite{kato2015open, hafeez2020insights}. 
The location is also published to the global planning node to 
calculate a navigation route to the 
destination~\cite{ros-global-planner}. The navigation route (step 4) 
and sensor fusion results (step 5) are both published to the local 
planning stage~\cite{ros-local-planner}, which constructs a local 
driving space cost map and generates and publishes vehicle 
trajectories to the vehicle's drive-by-wire system 
(step 6)~\cite{bertoluzzo2004drive, pan2006new}. 
Finally, the drive-by-wire system will send control messages to 
Electronic Control Units (ECUs) through the Controller Area Network 
(CAN) to maneuver the vehicle \cite{farsi1999overview}. 

Within the processing pipeline, the multi-tenant DNN inference 
serves as a foundation of the AV perception system. Multi-tenant DNN 
tasks run concurrently, merging their outputs based on timestamps 
to provide real-time updates on nearby moving objects, drivable 
areas, traffic lights/signs, etc.~\cite{yurtsever2020survey}.

\subsection{Motivation for Predictable Perception}

The safety of AVs hinges on real-time and accurate perception 
of their environment, necessitating predictability in both temporal 
and functional aspects~\cite{lin2018architectural}. Temporal 
predictability ensures the perception process is completed within a 
anticipated time frame, allowing other modules to respond 
appropriately. Functional predictability requires the 
perception module to provide accurate and comprehensive coverage 
of the dynamic traffic environment.

We address these challenges by proposing a novel approach 
that \textit{reduces the number of image frames processed to improve 
timing performance without sacrificing accuracy.} By dynamically 
adjusting critical frames and ROIs using temporal locality, 
we improve timing and functional predictability.

\section{Empirical Studies}
\label{sec:insights}

% \subsection{Opportunities for Predictable DNN inference for AV perception}

% To guarantee the AV's safety, the perception should adapt to a complex 
% environment which varies with time and location. 
% Unlike other work 
% focusing on model compression or stationary frame dropping~\cite{han2015deep, wu2020integer, yan2018deep}, 
% we explore from a very different perspective: can we dynamically adjust critical 
% frames and regions of interest to achieve timeliness and accuracy 
% simultaneously? 
This section provides some of the insights gained from our 
empirical studies of temporal locality and inference-time 
variations in the AV's perception.

% \vspace{0.5em}
% \noindent\textbf{C1: Environmental-aware Dynamic ROI} 
\subsection{Revisiting Temporal Locality}

Although a camera usually produces frames at 30 FPS, 
it is not necessary to process all of them owing to 
the high similarities between the consecutive frames. 
The SOTA approaches focus on leveraging the locality of pixels 
for real-time DNN inference~\cite{xu2018deepcache, li2020towards}. 
However, these similarity-based approaches ignore 
environmental information, thus limiting their capability.  

\begin{figure}[!htp]
	\centering
	\includegraphics[width=\columnwidth]{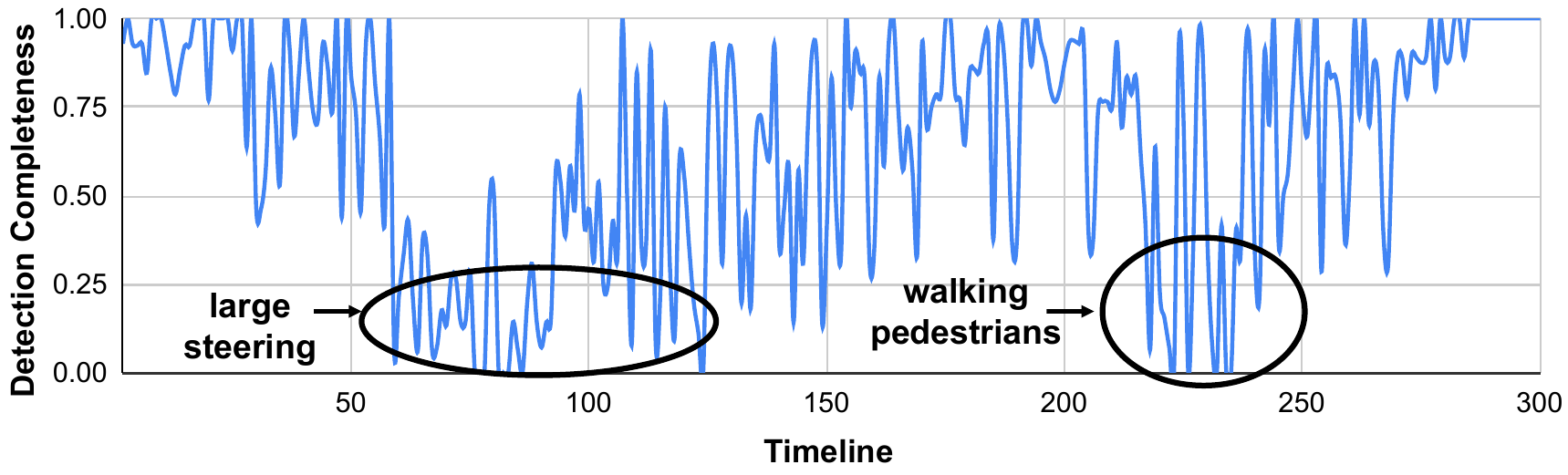}
        \vspace{-2em}
	\caption{The timing analysis of the detection completeness ratio for each frame when FPS = 5.}
	\label{fig:fps5-timeline}
\end{figure}

% define critical frames

\vspace{0.25em}
\noindent\textbf{Environment-aware critical frames.} We argue that 
the selection of critical frames should be scenario-aware, 
i.e., the number of critical frames should vary with the 
underlying traffic scenario. 
Fig.~\ref{fig:fps5-timeline} provides a timing analysis for the 
detection completeness relative to a 30 FPS speed when the FPS 
is reduced to 5. Detection completeness is calculated as the 
percentage of objects being detected from selected critical 
frames relative to the case of processing all frames. 
From the timeline, we can observe two periods of low detection 
completeness. 
With some driving context information from the image stream, 
we found the large steering control during the first period
caused higher pixel changes in consecutive frames. 
Nearby walking pedestrians caused low detection
completeness during the second period. 
The movement of pedestrians makes the intersection over 
union (IoU) lower than 0.5 in the original image. 
In both cases, it could cause safety problems if the 
detection misses occur. Therefore,  critical frames must 
be selected in an environmental-aware manner.

\begin{figure}[!htp]
	\centering
	\includegraphics[width=\columnwidth]{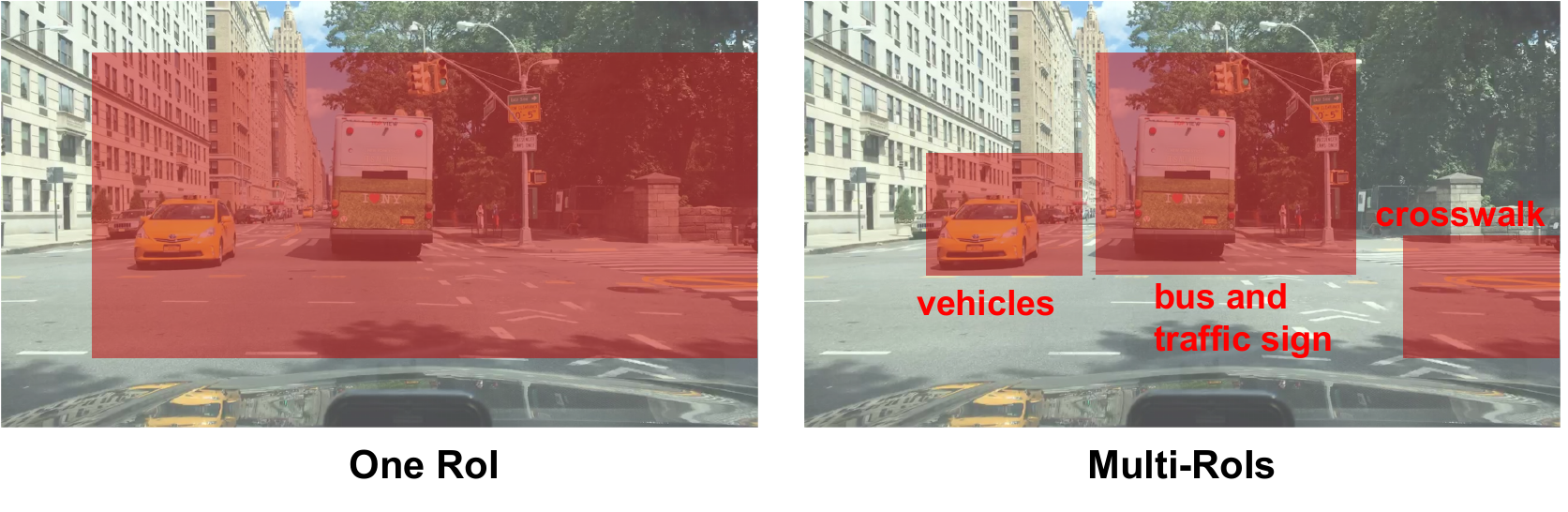}
        \vspace{-2em}
	\caption{An example of environment-aware dynamic ROIs.}
	\label{fig:RoI-example}
\end{figure}

\vspace{0.25em}
\noindent\textbf{Environment-aware dynamic ROIs.} We argue the ROI
must also be environment-aware for both critical and non-critical 
frames~\cite{kang2022dnn}. An ROI describes the critical region 
that must be detected both timelily and accurately. 
The SOTA approaches mainly consider one ROI on a single frame, 
usually requiring the ROI to cover all the areas of interest. 
In contrast, we found that multiple ROIs can be more efficient 
than a single ROI in terms of reducing input size. 
Fig.~\ref{fig:RoI-example} shows an example of one and 
multiple ROIs in a single frame. One can observe that the ego 
vehicle needs to track the vehicle on the left, the bus in front, 
traffic lights and signs, and pedestrians crosswalk on the right. 
The left image shows the use of one big ROI to cover all these 
areas, while the right image uses three ROIs to cover the vehicle, 
the bus and traffic sign, and the crosswalk. 
Both ensure comprehensive coverage of essential areas. 

% Both cases have vehicles running from opposite directions, while the difference is the right has a yellow lane to separate the traffic flowing in opposite directions. We can observe that moving objects (vehicles) and traffic signs are shown in limited areas for both cases. Similarly, the determination of ROI when the vehicle is stopping at a traffic intersection or moving should also be different. The potential moving objects (vehicles, pedestrians, bicycles), vehicle speed, road levels (highway, downtown), and environmental conditions (daytime, night, rain, snowing) should all be considered to determine the dynamic ROIs for each frame.

% Modeling and detection of the best ROI for a dynamic environment

% Why? 1. statically dropping frames will lose accuracy; 2. not all areas for critical frames are equally important

\noindent\textbf{Insight 1:} \textit{Critical frames and ROIs 
vary with the environment and needs to be determined dynamically. 
Using multiple ROIs can reduce the size of critical input
more effectively than a single ROI.}

% \subsection{Non-linear Relationship between ROI Size and Flops}

\subsection{Revisiting Inference Time Variations}

The use of dynamic critical frames and ROIs reduces the size of 
input to the DNN inference, but also increases the
inference-time variation. 
The trained DNN models with two-stage-based structures are shown to 
have higher inference-time variations than one-stage-based 
structures~\cite{liu2023understanding, liuprophet, li2020towards}. 
Note, however, that we focus on the inference-time variations 
caused by variable ROI size and multi-tenant DNNs.

\begin{figure}[!htp]
	\centering
	\includegraphics[width=\columnwidth]{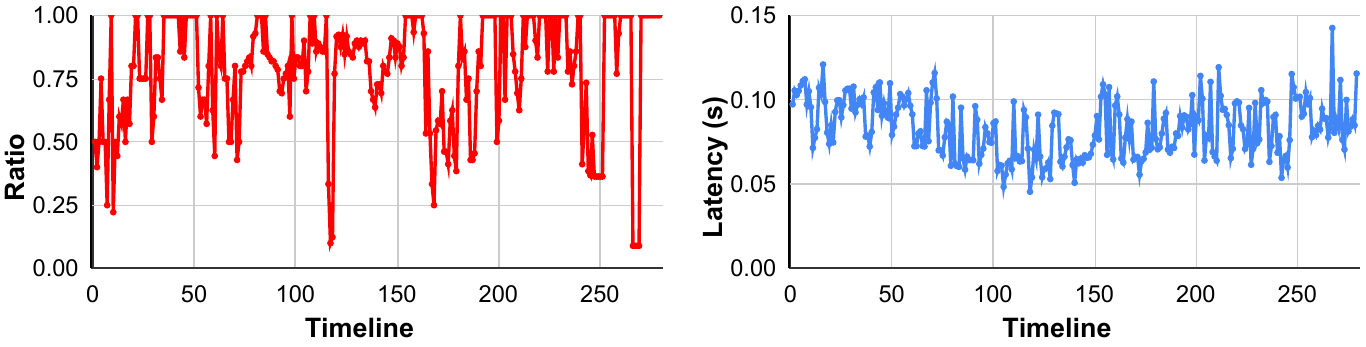}
        \vspace{-2.1em}
	\caption{The area ratio for consecutive frames and 
   the corresponding inference time using Faster R-CNN.}
	\label{fig:ratio-latency}
\end{figure}

% main sources: 1. model structure (two-stage model); 2. input size (dynamic ROI);

\vspace{0.25em}
\noindent\textbf{Image size and time variations.} In general, 
the ROI needs to cover all the critical objects, so we define ROI 
as the minimum cover of all the detected objects in each frame. 
By applying this definition to the BDD100K dataset, we 
calculate the ratio of ROI area to the original image size. 
Fig.~\ref{fig:ratio-latency} shows the timeline of the 
ratio and the inference latency when the cropped images are
applied to the Faster R-CNN model. One can observe that the 
ratio changes dynamically and is mostly less than 1, making 
it unnecessary to process the entire frame. 
Moreover, dynamic ROIs contribute to the inference-time 
variations of the Faster R-CNN model. 
% By applying the ROI cropped images and the original images to 
% DNN models (Faster R-CNN~\cite{ren2015faster}, 
% DNLNet~\cite{yin2020disentangled}, 
% Deeplabv3~\cite{chen2017rethinking}), we analyze the mean latency, 
% standard deviation, coefficient of variations, and range. 
% From Table~\ref{tab:roi-wo-models}, we can observe the mean 
% latency is reduced with consideration of dynamic ROIs at the 
% expense of increased inference-time variations.

% \usepackage{tabularray}
% \begin{table}
% \centering
% \caption{Inference-time statistics with and without ROIs.}
% \vspace{0.5em}
% \label{tab:roi-wo-models}
% \resizebox{.8\columnwidth}{!}{%
% \begin{tblr}{
%   cells = {c},
%   cell{2}{1} = {r=2}{},
%   cell{4}{1} = {r=2}{},
%   cell{6}{1} = {r=2}{},
%   hline{1,8} = {-}{0.08em},
%   hline{2,4,6} = {-}{},
% }
% \textbf{Models}       & \textbf{Latency}     & \textbf{mean (ms)}   & \textbf{sdv}  & \textbf{$C_v$}   & \textbf{range (ms)}  \\
% Faster R-CNN & w/o ROI & 104.09 & 0.01 & 0.07 & 65.05  \\
%              & w/ ROI    & \textbf{87.55}  & 0.01 & \textbf{0.16} & \textbf{104.16} \\
% DNLNet       & w/o ROI & 121.15 & 0.01 & 0.06 & 73.46  \\
%              & w/ ROI    & \textbf{89.65}  & 0.02 & \textbf{0.23} & \textbf{105.61} \\
% Deeplabv3    & w/o ROI & 145.87 & 0.01 & 0.10 & 105.90 \\
%              & w/ ROI    & \textbf{113.36} & 0.03 & \textbf{0.26} & \textbf{155.58} 
% \end{tblr}
% }
% \vspace{-0.5em}
% \end{table}

\begin{figure}[!htp]
	\centering
	\includegraphics[width=\columnwidth]{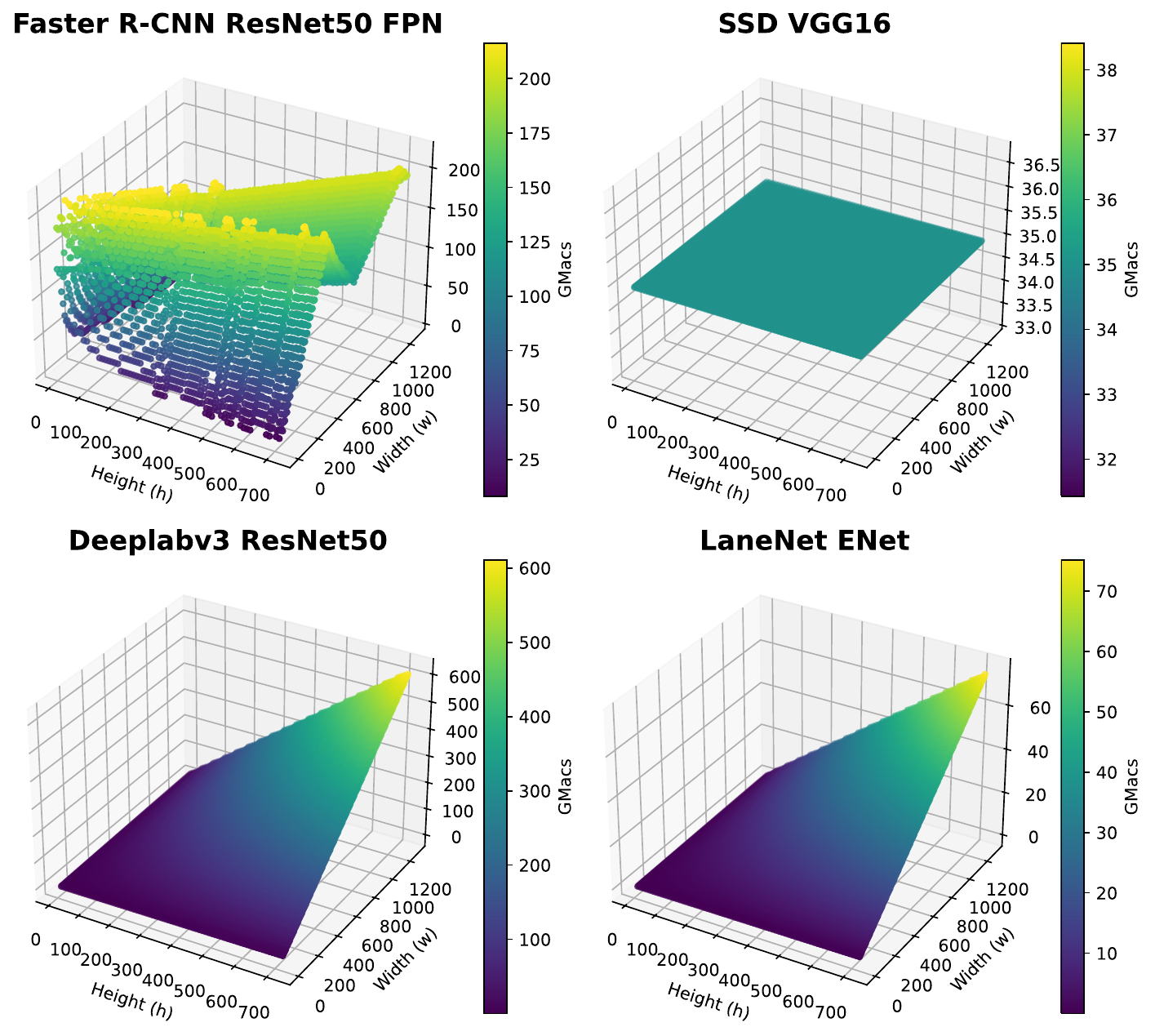}
        \vspace{-2em}
	\caption{GMACs of different image sizes (Height, Width) for DNN 
    models in the perception.}
	\label{fig:flops-hw}
\end{figure}

To better understand the relationship between image size and 
inference-time variations, we generated different image sizes:
width ranging from 10 to 720 and height ranging from 10 to 1280, 
both with a step size of 10. By applying these images of different 
sizes to the model for inference, we collect the 
Multiply-accumulate operations (MACs) which dominate 
the inference time. 
Four DNN models---Faster R-CNN~\cite{ren2015faster}, 
SSD~\cite{liu2016ssd}, Deeplabv3~\cite{chen2017rethinking}, 
LaneNet~\cite{neven2018towards}---in AV perception are tested 
with variable input-image sizes.
Both Deeplabv3 and LaneNet show a normal trend of large image 
sizes having higher Giga Multiply-Add Operations (GMACs).
Fig.~\ref{fig:flops-hw} shows the scatter plot of GMACs 
for Faster R-CNN and SSD. 
We can observe that Faster R-CNN shows a special GMACs peak with 
some small images while SSD shows static GMACs for all image sizes. 
We find the root cause of this difference from layer-wise GMACs 
and input/output sizes, i.e., pre-processing images causes 
this difference.
All two-stage-based models, like Faster R-CNN, are designed to take 
variable size inputs and the images go through 
variable ranges~\cite{ren2015faster}.
This resizing makes some small images have large MACs. 
One-stage model like SSD takes static input so all the images are resized to a fixed size. In contrast, segmentation models 
like Deeplabv3 and LaneNet do not resize the images. 
To determine the ROIs with minimum MACs, the 
resizing during the pre-processing of DNN models needs to 
be accounted for.

\begin{figure}[!htp]
	\centering
	\includegraphics[width=\columnwidth]{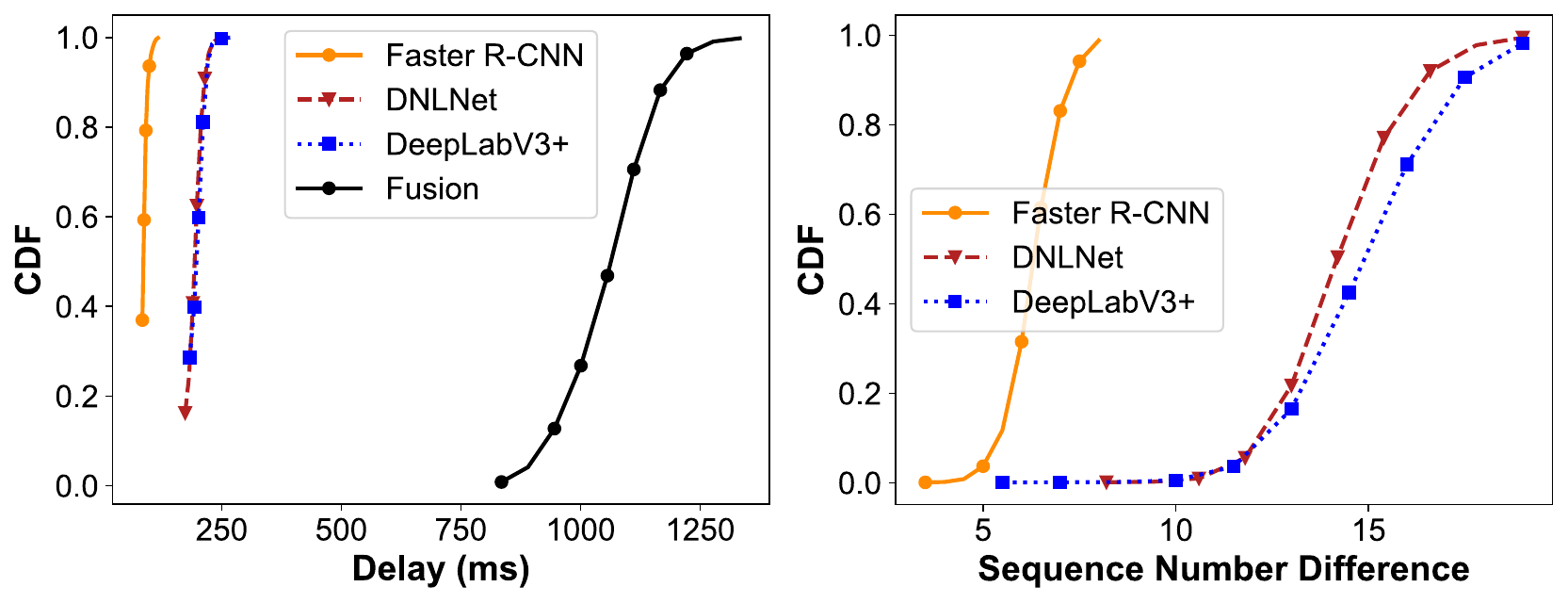}
        \vspace{-2em}
	\caption{The CDF of DNN inference delay, fusion delay, 
 and sequence number difference}.
 %CCCC What's sequence number delay?
 % sequence number is a unique number to images when publishing through ROS, sequence number delay means the sequence number difference between the newest frame and the frame finishing inference
 %CCCC A strange term
 % how about sequence number difference, it actually reflects the waiting time in the queue
	\label{fig:delay-fusion}
\end{figure}

\vspace{0.25em}
\noindent\textbf{Unpredictability of Multi-tenant DNNs Inference Time.} 
To understand the inference-time variations in multi-tenant DNNs, three 
DNN models are deployed onto three GPU cards (Faster R-CNN, DNLNet, and 
Deeplabv3+), and \texttt{message\_filter} in an ROS is leveraged to 
merge results in a time window~\cite{quigley2009ros,message-filter}. 
A group of over 4800 images from the BDD100K dataset are published 
as the ROS image streams at 30 FPS. 
Each image is assigned a unique sequence number, starting from 0. 
To process these images, three DNN tasks are initialized, waiting 
for ROS image data. These tasks are triggered by callback functions ---
where a function is passed as an argument into another function --- 
to be invoked or executed within the outer function. This ensures 
the tasks respond immediately to incoming images for processing. 
Fig.~\ref{fig:delay-fusion} shows the result for the inference 
delay of each task and fusion delay. It also includes the 
cumulative distribution function (CDF) of frame sequence 
number difference when entering callback functions.
The frame sequence number difference reflects the queuing delay 
of the frame entering the callback function. 
We can observe that the fusion delay is much 
higher than the combination of the inference-time variation for 
each single DNN. The fusion delay 
increases because the ROS subscriber uses multiple threads to 
handle incoming ROS images. Consequently, some threads process 
outdated frames from the message queue. This observation is 
corroborated by the CDF of the frame sequence number difference, 
where a notable portion of delay falls within the 10--15. 
Given the execution on GPU is not preemptable~\cite{cuda-driver}, 
the coordination of multi-tenant 
DNNs should be ROIs-aware.

\noindent\textbf{Insight 2:} \textit{Dynamic input size and out-of-date 
frames are the root cause of inference-time variations in the AV's 
perception. There is a non-linear relationship between input size and 
GMACs for DNN models. Coordination of multi-tenant DNNs inference 
should be ROI-aware.}

\section{System Design}
\label{sec:design}

%Predictability is essential for AV safety. 
We present the design of \name, including an ROI generator 
to select critical frames with dynamic ROIs for the AV's 
perception, a predictor to cache and predict the detection 
of objects, and a coordinator to determine/identify critical frames
and ROIs for each DNN inference task.
%CCCCC What do you mean by "dispatch the DNN inference"?
% It determines whether to skip frames or apply keyframe detection with ROIs to each DNN inference task.

\subsection{Overview}

\begin{figure*}[!htp]
	\centering
	\includegraphics[width=.9\textwidth]{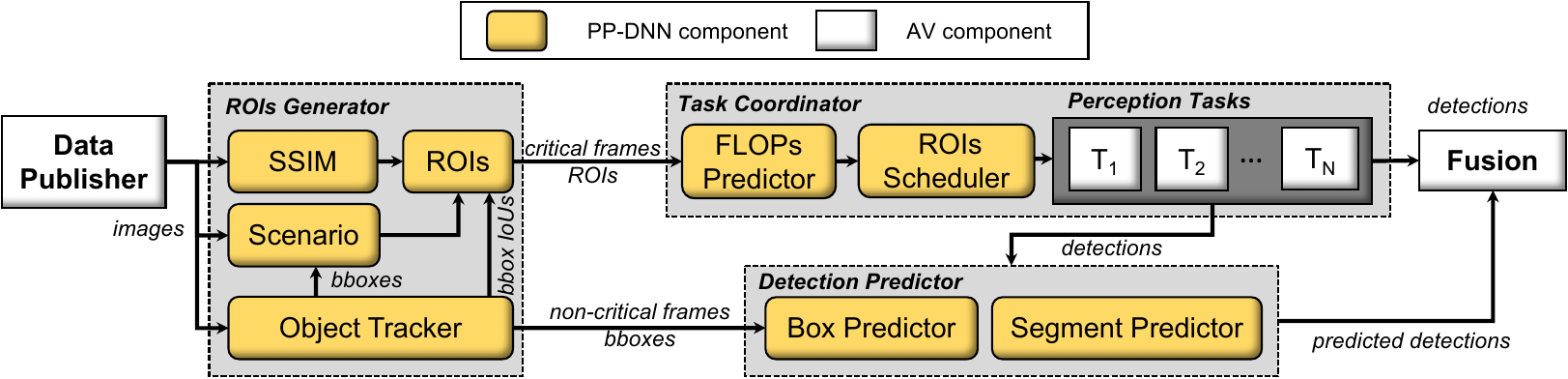}
        \vspace{-1em}
	\caption{An overview of \name.}
	\label{fig:system-overview}
\end{figure*}

Fig.~\ref{fig:system-overview} provides an overview of \name. 
\name\ supports the predictability of the AV's perception with 
three functional components/modules. 
First, the ROIs generator determines/identifies critical frames from 
the image stream by leveraging the pixel-level SSIM index and box-level 
tracking results. Depending on the location of detected objects and 
their environmental context, the generator produces a set of ROIs for 
each task. This set encompasses both single and multiple ROIs. 
Second, the selected critical frames and their associated ROIs are 
forwarded to the task coordinator. The expected FLOPs for each ROI are 
estimated using the FLOPs predictor. The frame scheduler then selects 
the ultimate critical frames and their respective ROIs for each task, 
prioritizing those with the lowest FLOPs. Third, the detection predictor 
publishes the detection results for non-critical frames to the fusion 
node. This predictor comprises both a box predictor and a segmentation 
predictor. The former retains all the tracking and detected bounding boxes 
associated with critical frames. Meanwhile, the segmentation predictor, 
informed by the ROIs and the box predictor's outputs, forecasts 
the current state of semantic segmentation. These predictions are 
updated at the image stream's frame rate, ensuring detection results
at a high frequency. These results are then published as ROS messages.

\subsection{Technical Challenges and Contributions}

To support a predictable perception pipeline, \name\ addresses 
the following challenges:

% \vspace{0.5em}
\noindent\textbf{C1:} \textit{How to model and detect ROIs in dynamic 
environments?} The ROIs are expected to 
provide minimum coverage of all the areas critical to AV safety. 
The ROIs should also be adapted to the dynamic environments. 
The pixel-wise similarity is commonly used to filter critical 
frames. In \name, we check the pixel- and bounding-box-level 
similarity and the traffic scenario in real time.

\noindent\textbf{C2:} \textit{How to choose ROIs to reduce 
inference-time variations?} Dynamically changing input size 
causes the DNN inference time to vary. \name\ uses the
relationship between MACs and image size to determine ROIs 
for all perception tasks.

\noindent\textbf{C3:} \textit{How to predict the detection results for 
non-critical frames?} Compared to running inference on critical frames 
with ROIs, the detection results for non-critical frames are predicted 
based on previous critical-frame detection and temporal locality.

\noindent\textbf{C4:} \textit{How to coordinate multi-tenant DNNs 
inference to reduce fusion-time variations?} In \name, the coordinator 
determines the ROIs and the critical frame execution order to reduce the fusion delay.

%% it means the execution order
%CCCC order of executing what?
% it means the critical frame execution order

\subsection{ROIs Generator}
\label{subsec:roi-generator}

% Since there is a high temporal locality of pixel distribution and objects in continuous frames, selecting some representative frames for environment perception is possible. Here the critical frame is defined as the frame with more representative features or pixels than the other frames~\cite{yan2018deep,rochan2019video}. 

% To guarantee the effectiveness of critical frames, selecting critical frames that can capture all the traffic environment changes becomes our primary objective. The critical frame selection in \name is based on several factors: pixel-level SSIM, bounding box tracking, traffic scenario, and task deadline.

To guarantee functional predictability, the choice of ROIs should be 
environment-aware and provide complete coverage of the moving objects 
and traffic segments. The ROIs generator produces critical frames 
of ROIs based on the pixel-level SSIM, bounding boxes tracking, 
and traffic scenarios.

\vspace{0.25em}
\noindent\textbf{Pixel-level SSIM.} SSIM reflects the structural 
similarity of pixels of images. To ensure the SSIM index is sensitive 
to object-level pixel changes, we first resize the image to a 
25$\times$25 greyscale image using a high-quality Lanczos 
filter~\cite{clark2015pillow}. The SSIM is calculated based 
on~\cite{wang2004image}. A sliding window of size 11$\times$11 with 
step size 1 is used to get patches from both images. 
For each patch, the SSIM is calculated using Eq.~(\ref{equation:ssim}), 
while the final SSIM of the image is the average SSIM of all the 
patches. When 
calculating the mean and variance of each patch, a Gaussian 
convolution kernel with a variance of 1.5 is used as a weighted 
average~\cite{wang2004image}. If the SSIM with the previous frame 
is less than a threshold, the current frame will be critical.

% \vspace{-2em}
\begin{equation}
\label{equation:ssim}
\operatorname{SSIM}(x, y)=\frac{\left(2 \mu_{x} 
\mu_{y}+c_{1}\right)\left(2 \sigma_{x y}+c_{2}\right)}
{\left(\mu_{x}^{2}+\mu_{y}^{2}+c_{1}\right)\left(\sigma_{x}^{2}+
\sigma_{y}^{2}+c_{2}\right)}
\end{equation}

%% a figure to show that tracking is pretty accurate in a short time, less than 1s
\vspace{0.25em}
% \vspace{-0.5em}
\noindent\textbf{Bounding boxes IOU.} Bounding box-level IOU 
reflects the locality of the detection results. We implement a 
light-weight object tracker based on DeepSort with YOLOv4-tiny to 
predict and update the ID and bounding boxes of moving objects for 
each frame~\cite{wojke2017simple,wojke2018deep,bochkovskiy2020yolov4}. 
Given the pixel- and box-level temporal locality for high FPS 
consecutive frames, the tracking can be accurate for most of the
time~\cite{li2020towards}. The output of the tracker includes 
all missed, tracked, and detected objects. For all tracked objects, 
bounding box IOU is calculated to show the box-level similarity of 
two consecutive frames. 
If either the number of missed objects or the average bounding box 
IOU for tracked objects is less than a preset threshold, 
the current frame will be critical. 

\vspace{0.25em}
\noindent\textbf{IOUs.} After tracking, a bounding box will be 
generated for each tracked object. 
Suppose $n$ is the sequence number of the current frame and $M$ is 
a set that contains all the tracked objects. 
Based on the tracking objects, we can generate two types of ROIs for 
each task from the ROIs generator: one or multiple ROIs. 
Fig.~\ref{fig:RoI-example} shows an example for both one 
and multiple ROIs choices. 
Multi-ROIs contain all the objects' bounding boxes together 
according to Eq.~(\ref{equ:multi-roi}), while a single ROI uses 
one rectangular box to cover all the tracked bounding boxes. 
Eq.~(\ref{equ:one-roi}) shows how to calculate by finding the top 
left corner with the width and the height.
% Multi-ROIs leverage the 
% box velocity to the current frame's box. 
% As Eq.~(\ref{equ:multi-roi}) shows, the box velocity in height 
% and width is added to the bounding box for each object.
% Fig.~\ref{fig:track-num} shows the CDF for the number of missed, tracked, and detected objects for each frame using YOLOv4-tiny and YOLOv4 as the detection model. We can observe that for both YOLOv4-tiny and YOLOv4 cases, the number of tracked objects aligns very well with the number of detected objects. The only difference is that YOLOv4 can detect more small objects than YOLOv4-tiny. However, a lightweight tracker is enough for calculating the box-level difference. We collected the number of missed objects and used the predicted bounding box positions to calculate the IOUs with the former frame. If the number of missed frames is larger than a threshold or the IOU is less, the frame will be marked as a critical frame.

% \vspace{-1em}
\begin{align}
    % \label{equ:box-velocity}
    % &V_{box}^{(j)} = \begin{aligned}[t]
    %     (x_{j}^{(n)}-x_{j}^{(n-1)}, y_{j}^{(n)}-y_{j}^{(n-1)}, \\
    %     h_{j}^{(n)}-h_{j}^{(n-1)}, w_{j}^{(n)}-w_{j}^{(n-1)})
    % \end{aligned} \\
    \label{equ:multi-roi}
    &\text{Multi-ROIs:} = \begin{aligned}[t]
        \{x_{j}^{(n)}, y_{j}^{(n)}, h_{j}^{(n)}, w_{j}^{(n)}\}, \forall j \in M
    \end{aligned} \\
    % \label{equ:multi-roi}
    % &\text{Multi-ROIs:} = \begin{aligned}[t]
    %     \{x_{j}^{(n)}, y_{j}^{(n)}, 2h_{j}^{(n)}-h_{j}^{(n-1)}, \\ 2w_{j}^{(n)}-w_{j}^{(n-1)}\}, \forall j \in M
    % \end{aligned} \\
    \label{equ:one-roi}
    &\text{One-ROI:} = \begin{aligned}[t]
        \{\min\limits_{\forall j \in M}x_{j}^{(n)}, \min\limits_{\forall j \in M}y_{j}^{(n)}, \max\limits_{\forall j \in M}h_{j}^{(n)}, \max\limits_{\forall j \in M}w_{j}^{(n)}\}
    \end{aligned}
\end{align}

% \begin{align}
%     \label{equ:box-velocity}
%     V_{box}^{(j)} &= (x{_{j}^{(n)}}-x{_{j}^{(n-1)}}, y{_{j}^{(n)}}-y{_{j}^{(n-1)}}, h{_{j}^{(n)}}-h{_{j}^{(n-1)}}, w{_{j}^{(n)}}-w{_{j}^{(n-1)})} \\
%     \label{equ:one-roi}
%     \text{One ROI:} &= (x{_{j}^{(n)}}-x{_{j}^{(n-1)}}, y{_{j}^{(n)}}-y{_{j}^{(n-1)}}, h{_{j}^{(n)}}-h{_{j}^{(n-1)}}, w{_{j}^{(n)}}-w{_{j}^{(n-1)})
% \end{align}

% \begin{equation}
%     \label{equ:box-velocity}
%     \resizebox{\columnwidth}{!}{%
%     $ V_{box}^{(j)} = (x{_{j}^{(n)}}-x{_{j}^{(n-1)}}, y{_{j}^{(n)}}-y{_{j}^{(n-1)}}, h{_{j}^{(n)}}-h{_{j}^{(n-1)}}, w{_{j}^{(n)}}-w{_{j}^{(n-1)})} $
%     }
% \end{equation}

% \begin{equation}
%     \label{equ:one-roi}
%     \resizebox{\columnwidth}{!}{%
%     One ROI: $ (x{_{j}^{(n)}}-x{_{j}^{(n-1)}}, y{_{j}^{(n)}}-y{_{j}^{(n-1)}}, h{_{j}^{(n)}}-h{_{j}^{(n-1)}}, w{_{j}^{(n)}}-w{_{j}^{(n-1)})} $
%     }
% \end{equation}

% \begin{figure}[!htp]
% 	\centering
% 	\includegraphics[width=\columnwidth]{figures/insights/cdf-num-yolov4-tiny.pdf}
%         \vspace{-2em}
% 	\caption{The CDF of missed, tracked performance for YOLOv4-tiny and YOLOv4.}
% 	\label{fig:track-num}
% \end{figure} 

\vspace{0.25em}
\noindent\textbf{Scenario \& time interval awareness.} Another critical 
factor that could impact AV safety is the traffic scenario. 
Based on the detection results for objects, the traffic scenario is 
determined based on the percentage and class of moving objects and 
stationary objects. The target moving objects include all types of 
vehicles (cars, buses, trucks, motorcycles, etc.), pedestrians, and 
bicycles. Stationary objects include traffic signs and traffic lights. 
The ratio of each class will be calculated and monitored. If the 
stationary object ratio is close to 0, the choice of ROIs will be 
more focused on dynamic objects' bounding boxes. 
Otherwise, a dedicated ROI will be assigned to keep track of 
traffic lights/signs. For moving objects, if the pedestrians' 
ratio is $>0.5$, a ROI with 
complete object coverage will be used.

The time interval is another factor for critical frame selection. 
Although there is a minor difference between consecutive frames, 
the accumulation of slight differences could be large. 
Therefore, we monitor the time interval between critical frames and 
set a timer with the deadline. 
If the time interval between the current time and the last critical 
frame exceeds the deadline, the current frame will be 
a critical one. 

\begin{algorithm}
\caption{Critical Frame Selection Process}
\label{alg:critical-frame-selection}
\begin{algorithmic}[1]
\small
\State \textbf{Input:} Current frame $F_{current}$, Previous frame $F_{previous}$, Critical frame interval $T_{interval}$, Traffic scenarios $S_{traffic}$, SSIM threshold $\tau_{SSIM} = 0.95$
\State \textbf{Output:} Set of critical frames $C$

\State $C \gets \emptyset$ \Comment{Initialize the set of critical frames}

\For{each frame $F_{current}$}
    \If{$T_{interval} > deadline$ \textbf{or} $F_{current} \in S_{traffic}$}
        \State $C \gets C \cup \{F_{current}\}$ \Comment{Mark frame as critical}
    \Else
        \State $SSIM \gets \textbf{SSIM}(F_{current}, F_{previous})$
        \If{$SSIM < \tau_{SSIM}$}
            \State $C \gets C \cup \{F_{current}\}$ \Comment{Mark frame as critical}
        \Else
            \State $bbox\_pred$, $obj\_miss \gets \textbf{Tracker}(F_{current})$
            \If{$obj\_miss > 1$}
                \State $C \gets C \cup \{F_{current}\}$ \Comment{Mark frame as critical}
            \EndIf
        \EndIf
    \EndIf
    
    \If{$F_{current} \in C$} 
        \State \text{Generate one-ROI / multi-ROIs}
    \EndIf
\EndFor

\State \Return $C$
\end{algorithmic}
\end{algorithm}

% \vspace{0.25em}
% \noindent\textbf{Critical frame selection process.} In summary, 
% the selection of critical frames and ROIs consists of the following steps:
% % \vspace{-0.5em}
% \begin{enumerate}[noitemsep,topsep=0pt]
%     \item Check the critical-frame time interval to see if the interval 
%        has passed the deadline, or if the current frame belongs to a 
%        specific traffic scenario. If yes, then this frame will be 
%        critical, else go to the next step;
%     \item Calculate the SSIM of the current frame with the previous 
%        frame. If the SSIM is less than 0.95, mark it as a critical frame, else go to the next step.
%     \item Feed the current frame into the tracker to get predicted 
%       bounding boxes and missed and tracked objects. If there is 
%       more than one missed tracking object, the frame will be 
%       marked as a critical frame and go to the next step.
%     \item If the current frame is critical, then generate 
%       both one-ROI and multi-ROIs based on the ROI model. 
%       Otherwise, move to the next frame.
% \end{enumerate}

\subsection{Detection Predictor}
\label{subsec:detect-predict}

In \name, a detection predictor is designed to temporally predict 
the bounding boxes and semantics for updating the detection 
results with high FPS. The predictor includes two parts: 
(i) box predictor, which predicts bounding boxes based on 
tracking results, and (ii) a segmentation predictor, which predicts 
the segmentation's current status and published as ROS messages.

\vspace{0.25em}
\noindent\textbf{Box predictor.} Based on the analysis of the 
fusion of multi-tenant DNN inference, the "wooden barrel effect`` 
slows down the fusion process. The detection process is the DNN 
inference which is time-consuming while the fusion node waits for 
the results for message synchronization. Therefore, \name\ addresses 
this challenge by decoupling the detection process with the results 
publishing process. Here we store separate detection queues for 
three tasks. For object detection, bounding boxes, scores, and 
timestamps are saved. For semantic segmentation and lane detection, 
segmentation images with pixels marked as segmentation labels are 
saved. The queue length is maintained to include 10 critical frame 
detections in the queue. The history of results will be deleted when 
new critical frames are added to the queue. The object 
detection prediction is mainly based on the tracker's output. 
For each tracked object, we calculate the maximum IoUs with the cached 
object detection results. If IoU is larger than 0.5, the bounding box 
will be updated with the tracking results. If the IoU is less than 0.1, 
the cached bounding box will be kept while another bounding box is
based on the tracker. The cached bounding box will be updated when 
a new critical frame completes the inference.

\begin{figure}[!htp]
	\centering
	\includegraphics[width=\columnwidth]{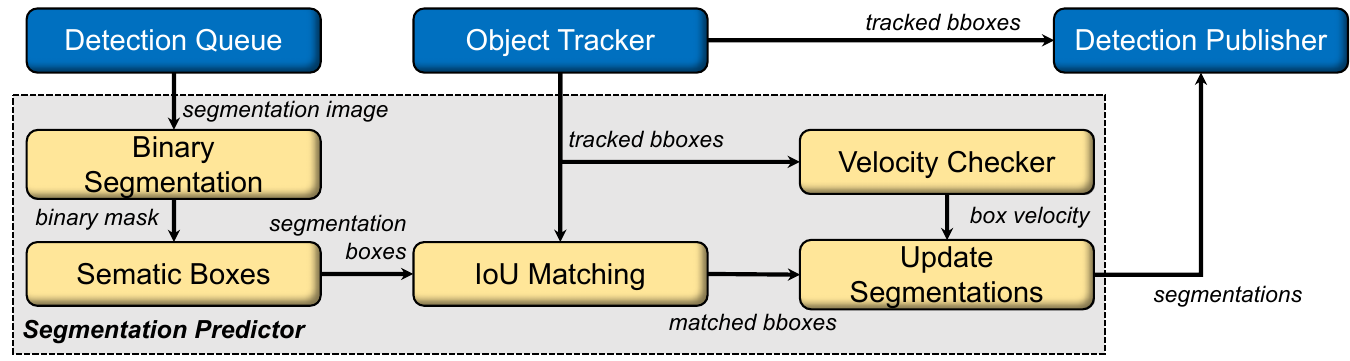}
        \vspace{-2em}
	\caption{The design of the segmentation predictor in the 
      predictor module.}
	\label{fig:segmentation-predictor}
\end{figure}

\vspace{0.25em}
\noindent\textbf{Segmentation predictor.} The segmentation predictor 
is responsive to predict the updated pixel semantics.  
Since both lane detection and semantic segmentation are updating 
segmentation images, we propose a velocity-based prediction for 
updating segmentation pixels. Fig.~\ref{fig:segmentation-predictor} 
shows the design of the segmentation predictor.

For segmentation images cached by the detection queue, the 
segmentation predictor first converts it to a binary segmentation 
image where segmentation will be one while the background will be 0. 
Next, segmentation bounding box coordinates will be extracted from 
the binary segmentation mask. Meanwhile, the tracker's output with 
tracked bounding boxes for objects is read, calculating the IOU with 
the segmentation bounding boxes. The two boxes are matched similarly 
if the IOU is larger than 0.5. The tracker also calculates the 
velocity, which includes the difference in the x-axis, y-axis, 
height, and width using the tracked bounding boxes' movement. 
Eq.~(\ref{equ:box-velocity}) shows the calculation of box velocity 
based on the coordinates of tracked boxes in continuous frames. 
For all matched boxes, the velocity is applied to update the 
segmentation pixels. Finally, the updated segmentation results will 
be sent to the detection publisher. For lane detection, we use the 
last critical frame's output as the updated lanes since we observe limited 
differences from the detection results of consecutive frames.

\vspace{-1em}
\begin{equation}
    \label{equ:box-velocity}
    \resizebox{\columnwidth}{!}{%
    $ V_{box} = (x{_{min}^{(n)}}-x{_{min}^{(n-1)}}, y{_{min}^{(n)}}-y{_{min}^{(n-1)}}, h^{(n)}-h^{(n-1)}, w^{(n)}-w^{(n-1)}) $
    }
\end{equation}

% The semantic update process contains the following three steps: 1. get velocity based on the movement of the object bounding box, and get the velocity set of the moving bounding box which has high matching and detection scores. 2. for cached previous semantic pixels, convert it to a binary semantic image and get bounding boxes based on the border (https://github.com/nikhilroxtomar/Semantic-Segmentation-Mask-to-Bounding-Box/blob/main/mask\_to\_bbox.py). Compare the semantic boxes with the tracker's bounding boxes to find the ones with IOU larger than 0.5 and the labels are the same, update the position of the semantic box using the corresponding velocity. Otherwise, no update;

\subsection{Task Coordinator}
\label{subsec:task-coordinate}

In \name, the task coordinator is designed to determine the 
critical frame and ROIs to reduce the fusion delay.

% \vspace{0.5em}
\noindent\textbf{FLOPs predictor.} With the understanding of 
the impact of image sizes on MACs for perception models in 
Section~\ref{sec:insights}, we formulate a model for 
predicting the total FLOPs for the perception pipeline. 
Eqs.~(\ref{equ:object-flops}),~(\ref{equ:lane-flops}), and
(\ref{equ:segment-flops}) show the FLOPs of object detection, 
lane detection, and segmentation for one given ROI box. 
We can observe that the FLOPs for the object detection model is 
segmented function with respect to the height/width ratio. 
This is mainly because the object detection model has a 
pre-processing stage that resizes the image into a range of size 
$[s_{min}, s_{max}]$. Besides, the ceiling is applied to the resized 
image to make all sizes divisible by $s_d$. The size range is 
[800, 1333] while the divisible size is 32. For segmentation tasks, 
since there is no resizing, the FLOPs are calculated directly. 

% Eq.~(\ref{equ:flops-all}) shows the total FLOPs for one frame with 
% one ROI or multi-ROIs. The coordinator can leverage this FLOP 
% prediction to control the frame sequence as well as the 
% execution progress of each task.

% \left \lceil \frac{s_{max}h}{w\cdot s_{d}} \right \rceil \cdot \left \lceil \frac{s_{max}}{s_{d}} \right \rceil \cdot s_{d}^{2}, 0 < \frac{h}{w} \leq \frac{s_{min}}{s_{max}}

% \\

% \left \lceil \frac{s_{min}}{s_{d}} \right \rceil \cdot \left \lceil \frac{s_{min}w}{h\cdot s_{d}} \right \rceil \cdot s_{d}^{2}, \frac{s_{min}}{s_{max}} \leq \frac{h}{w} \leq 1

% \\

% \left \lceil \frac{s_{min}h}{w\cdot s_{d}} \right \rceil \cdot \left \lceil \frac{s_{min}}{s_{d}} \right \rceil \cdot s_{d}^{2}, 1 \leq \frac{h}{w} \leq \frac{s_{max}}{s_{min}}

% \\

% \left \lceil \frac{s_{max}}{s_{d}} \right \rceil \cdot \left \lceil \frac{s_{max}w}{h\cdot s_{d}} \right \rceil \cdot s_{d}^{2}, \frac{s_{max}}{s_{min}} \leq \frac{h}{w}

% \vspace{-2em}
\begin{align}
\label{equ:object-flops}
f_{0}(h, w) &= 
\begin{cases} 
    a_{0}\left \lceil \frac{s_{max}h}{w\cdot s_{d}} \right \rceil \cdot \left \lceil \frac{s_{max}}{s_{d}} \right \rceil \cdot s_{d}^{2} & 0 < \frac{h}{w} \leq \frac{s_{min}}{s_{max}} \\
    a_{0}\left \lceil \frac{s_{min}}{s_{d}} \right \rceil \cdot \left \lceil \frac{s_{min}w}{h\cdot s_{d}} \right \rceil \cdot s_{d}^{2} & \frac{s_{min}}{s_{max}} \leq \frac{h}{w} \leq 1 \\
    a_{0}\left \lceil \frac{s_{min}h}{w\cdot s_{d}} \right \rceil \cdot \left \lceil \frac{s_{min}}{s_{d}} \right \rceil \cdot s_{d}^{2} & 1 \leq \frac{h}{w} \leq \frac{s_{max}}{s_{min}} \\
    a_{0}\left \lceil \frac{s_{max}}{s_{d}} \right \rceil \cdot \left \lceil \frac{s_{max}w}{h\cdot s_{d}} \right \rceil \cdot s_{d}^{2} & \frac{s_{max}}{s_{min}} \leq \frac{h}{w} \\
\end{cases} \\
\label{equ:lane-flops}
f_{1}(h, w) &= a_{1}hw \\
\label{equ:segment-flops}
f_{2}(h, w) &= a_{2}hw
% \label{equ:flops-all}
% F^{(n)} &= \sum_{\forall (h, w) \in ROIs}^{} f_{0} + f_{1} +f_{2}
\end{align}

\vspace{0.25em}
\noindent\textbf{ROIs scheduler.} Since different tasks could have 
different critical frames and ROIs, the ROI scheduler takes the 
FLOPs prediction results for each task and chooses 
$(h, w)$ to minimize the FLOPs for each frame. 
Multi-ROIs are packed as one batch for DNN inference, 
while ML libraries like Pytorch require all images in one batch 
to have the same size. Therefore, we apply the largest 
size in multi-ROIs to extract multi-ROIs before FLOPs prediction. 
The ROIs scheduler determines the choice of one-ROI or multi-ROIs 
based on the predicted FLOPs.

% Besides, the scheduler 
% also monitors the execution progress of multi-tenant tasks to 
% reduce fusion time variations by skipping non-critical and 
% out-of-date frames.

In the perception pipeline in Fig.~\ref{fig:e2e-pipeline}, each task—detection, 
segmentation, and fusion—subscribes to the camera stream and fires 
its callback as soon as a new frame arrives. A lightweight, distributed 
scheduler maintains a shared state that records the latest sequence numbers 
and deadlines, and dispatches critical frames according to three rules. 
First, if a task has already missed its per-frame deadline, the incoming 
frame is routed exclusively to that task. Second, if the traffic-scenario 
detector flags the frame as safety-critical (e.g., cut-in vehicle, pedestrian 
crossing), it is broadcast to all tasks. Third, for all other frames, 
the scheduler computes each task’s delay—the sequence-number gap between 
the candidate frame and the task’s most recent output—and drops the frame 
for any task whose delay exceeds a task-specific threshold. 
This threshold is derived from the static end-to-end deadline and 
the camera frame rate (e.g., six frames for a 200 ms budget at 30 FPS). 
Together with a prediction module that refreshes single-task outputs 
at the camera’s frame rate, this dispatcher bounds per-task latency and 
ensures timely multi-task fusion, enabling \name to meet the hard real-time 
requirement of autonomous driving.

\section{Implementation}
\label{sec:implementation}

To fill the gap for end-to-end perception testing and 
evaluation, we integrate the implementation of \name\ into 
an AV perception pipeline based on ROS in a GPU platform. 

\begin{figure}[!htp]
	\centering
	\includegraphics[width=\columnwidth]{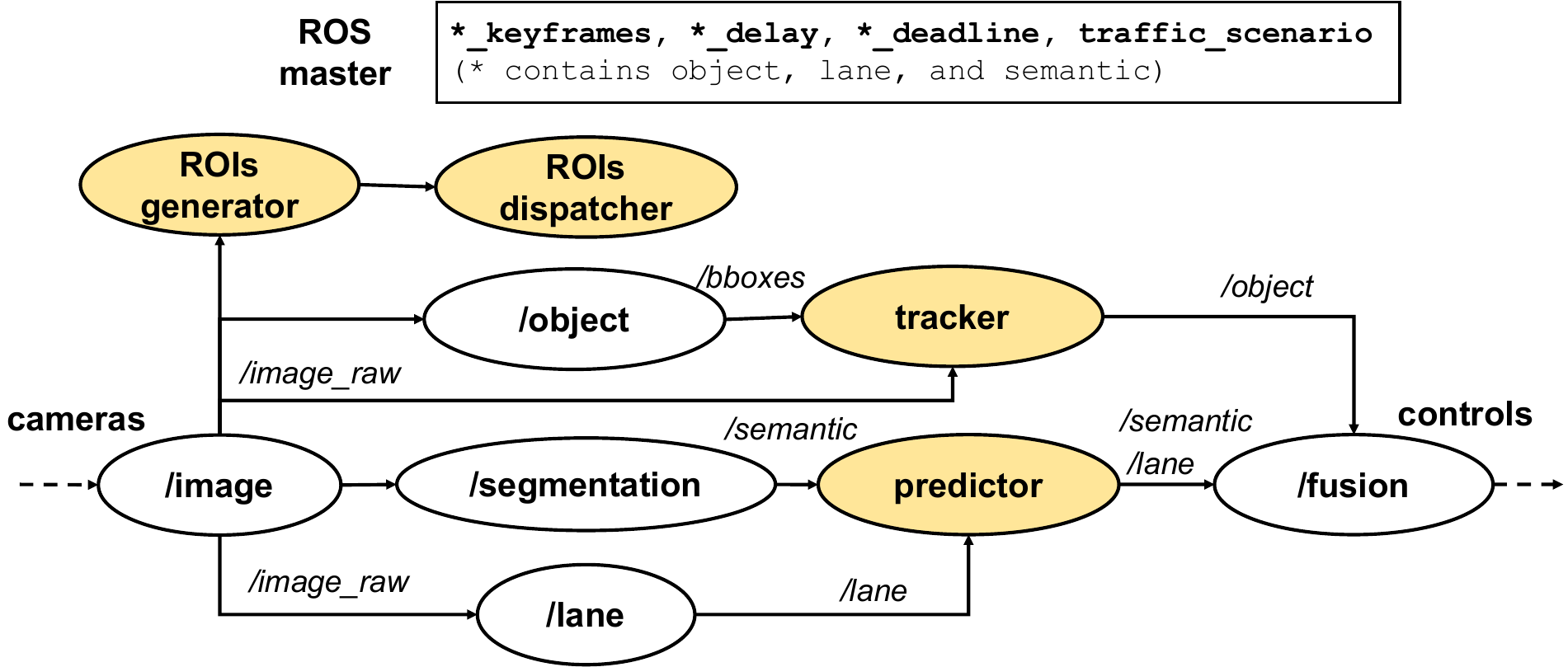}
        \vspace{-2em}
	\caption{The ROS implementation of the PP-DNN system.}
	\label{fig:ros-framework}
\end{figure}

\vspace{0.25em}
\noindent\textbf{ROS framework.} Based on the AV system 
in Fig.~\ref{fig:e2e-pipeline}, we develop a ROS framework for 
its perception system, as shown in Fig.~\ref{fig:ros-framework}. 
The ROS framework focuses on the DNN-based perception tasks. 
The pipeline starts with the \textit{/image} node, capturing and 
publishing images from the cameras. Three ROS nodes subscribe 
\textit{/image\_raw} messages and execute the DNN inference on 
the images, including object detection, lane detection, and 
semantic segmentation. The \textit{ROIs generator} node 
subscribes to the \textit{/image\_raw} topic, determines critical frames and ROIs, 
and sends to the \textit{ROIs dispatcher} node. The tracker and 
the predictor subscribe to the results from the three DNN tasks 
and publish predicted results to the \textit{/fusion} node.

Moreover, the critical frames on each task are dispatched based on the ROS 
parameters which are global variables managed by the ROS master. 
We define 10 ROS parameters for passing three tasks' critical frame 
sequence ID, delay, deadline, and traffic scenario.

\vspace{0.25em}
\noindent\textbf{Message synchronization.} In ROS, message 
synchronization is achieved by using the timestamp and sequence 
number/ID. The \textit{/fusion} node combines all the perception 
results of the same image frame. First, we 
have to generate a unique ID for each image frame. 
In the beginning, the \textit{/image} node attaches a timestamp and 
frame ID to each message it publishes. The timestamp and sequence ID 
of the incoming images will be used as the header's timestamp 
and sequence number of the new message like \textit{/object}, 
\textit{/semantic}, etc. 
With unique IDs on each image frame and detection results, we implement a 
\textit{message\_filter} with the Approximate Time Synchronizer to manage the 
fusion process~\cite{message-filter}. The approximate synchronizer sets 
queue size as 1000 and 300ms as the slop, which means the message with a time difference of less than 300ms is considered synchronized.

\vspace{0.25em}
\noindent\textbf{Detection completeness.} To show how good the selected 
critical frame' performance is in environmental perception, based on the Pseudo ground truth concept in~\cite{li2020towards}, we propose a 
detection completeness metric that calculates the number of detected 
objects in critical frames compared to the baseline, which processes all the frames. An object is detected if the detected bounding box has over 
0.5 IoU with the baseline's detection results. The pseudo-code for detection completeness calculation is shown in 
Algorithm~\ref{algorithm1}. The input contains the detection results for 
a group of frames for the keyframes and the baseline (process all the 
frames). First, identify the frame index $index\_key$ in the keyframes' 
results (Lines 3-5). Second, go through all the bounding boxes in that 
frame and calculate the max IoU with the keyframe's detection results 
(Lines 6-7). For each object, if the score is larger than 0.5, the number 
of objects will be increased by 1. If the max IoU is larger than 0.5 and 
the detected score in the keyframe is larger than 0.5, then the detected 
object is increased by 1. After checking all the frames, the detection 
completeness $d_c$ is calculated using $detected$ divided by $objects$.

\begin{algorithm}
\caption{Detection Completeness}
\label{algorithm1}
\begin{algorithmic}[1]
\small
\State \textbf{Input:} Keyframes' bounding boxes $D_k$, scores $S_k$, timestamps $t_k$; Offline detections' bounding boxes $D_a$, scores $S_a$, timestamps $t_a$
\State \textbf{Output:} Detection completeness $d_c$

\State $detected \gets 0$, $objects \gets 0$
\State $index\_key \gets 0$

\For{$index$, $frame$ \textbf{in} $D_a$}
    \If{$t_a[index] > t_k[index\_key]$}
        \State $index\_key \gets index\_key + 1$
    \EndIf
    \For{$box\_id$, $boxes$ \textbf{in} $frame$}
        \State $mIoU$, $nmax \gets$ \textbf{maxIoU}($D_k[index\_key]$, $boxes$)
        \If{$S_a[index][box\_id] > 0.5$}
            \State $objects \gets objects + 1$
            \If{$mIoU > 0.5$ \textbf{and} $S_k[index\_key][nmax] > 0.5$}
                \State $detected \gets detected + 1$
            \EndIf
        \EndIf
    \EndFor
\EndFor

\State $d_c \gets detected / objects$
\State \Return $d_c$
\end{algorithmic}
\end{algorithm}

\section{Evaluation}
\label{sec:evaluation}

We evaluate the effectiveness of \name\ in guaranteeing 
the predictability of AV perception through comprehensive 
experiments with an ablation study.
Our key findings are:
\begin{itemize}
  \item \name\ significantly improves multi-tenant DNN inference. 
    The detection predictor helps increase the number of fused 
    frames by up to 7.3$\times$. The frame dispatcher helps shorten the 
    fusion delay by $>$2.6$\times$ and fusion delay variations by $>$2.3$\times$. \name\ guarantees the timing 
    predictability of the perception pipeline. 
    (\S\ref{subsec:multi-dnn-fusion}).
  \item The detection completeness effectively captures the 
    streaming detection performance. 
    \nameS improves the completeness of object detection 
     by 75.4\%. It guarantees the functional predictability 
     for the perception pipeline. 
     (\S\ref{subsec:detection-completeness}).
  \item \name\ significantly improves latency and accuracy, 
    making up to a 98\% improvement in cost-effectiveness. 
    (\S\ref{subsec:cost-effectiveness}) while incurring 
    (acceptable) 20\% CPU and 39.3\% 
    GPU memory-consumption overheads.
\end{itemize}

\subsection{Experimental Setup}

We first present the setup used for the implementation 
and evaluation of \name. We choose a GPU platform as 
the computing device following general AV settings~\cite{AD-computer, aws-a10g} and use the BDD100K dataset and 
the nuScenes mini dataset as the input to the perception 
pipeline~\cite{yu2020bdd100k, caesar2020nuscenes}. 
The BDD100K dataset is composed of 2,629 image 
frames covering 12 traffic scenarios at 30 FPS. 
The nuScenes dataset includes 1,938 images covering 
10 traffic scenarios at 12 FPS.
These images are published as \texttt{ROS Image}.

\vspace{0.25em}
\noindent\textbf{Hardware and software setup.} 
The GPU platform has 8 Intel® Core™ i9-9940X CPUs with 
the highest frequency of 3.3GHz and 32 GB DDR4 CPU memory.
The platform has one NVIDIA A10G Tensor Core GPU cards with 
24GB GDDR6 GPU memory~\cite{aws-a10g}. It provides 31.2 teraFLOPS for FP32.
The libraries installed for ML-related applications 
include CUDA Driver 510.47.03, CUDA runtime 11.6, 
TensorFlow 1.15.2, torch v1.10.1, torchvision v0.11.2, 
cuDNN 8.3.2, OpenCV 4.2, etc. ROS Melodic is deployed 
as the communication middleware. 
Since it is essential to meet the required accuracy for AVs, 
all the DNN models are those trained from the BDD100K model zoo, 
and tested with full precision (FP32)~\cite{yu2020bdd100k}.
Faster R-CNN, DNLNet, and Deeplabv3+ are deployed for object 
detection, lane detection, and semantic segmentation
\cite{ren2015faster,yin2020disentangled,chen2017rethinking}. 
Approximate Time Synchronizer in ROS \textit{message\_filter} is 
deployed for sensor fusion~\cite{quigley2009ros,message-filter}.

% \usepackage{booktabs}
% \usepackage{graphicx}
\begin{table}[]
% \vspace{-0.5em}
\centering
\caption{The names and configurations of five testing cases.}
\label{tab:testing-case}
\resizebox{.8\columnwidth}{!}{%
\begin{tabular}{@{}cc@{}}
\toprule
\textbf{Testing Cases Name} & \textbf{Configurations}       \\ \midrule
Baseline                     & Base                    \\
\midrule
FD                     & Base + Dispatcher                   \\
FD+FG                      & Base + Dispatcher + ROI             \\
FD+DP                     & Base + Dispatcher + Predictor            \\
\midrule
\name\                      & Base + Dispatcher + ROI + Predictor \\ \bottomrule
\end{tabular}%
}
\vspace{-0.5em}
\end{table}

% Compare PP-DNN with baseline
% ablation study

\vspace{0.25em}
\noindent\textbf{Testing Cases.} We consider five testing cases 
to evaluate the impact of \name's modules on the fusion delay 
and detection completeness. FD follows the same design for 
multi-tenant DNNs coordination as Prophet~\cite{liuprophet}.
Table~\ref{tab:testing-case} shows the names and 
configurations of these five testing cases. 
In the base configuration, all the frames are published 
and processed in the ROS data stream system. 
We conduct ablation experiments on the impact of the 
frame dispatcher (FD), the critical frame ROI generator (FG), 
the detection predictor (DP), and \name. %separately. 

\vspace{0.25em}
\noindent\textbf{Metrics.} We primarily focus on three metrics: latency, accuracy, and 
cost-effectiveness. In terms of latency, we evaluate both the single DNN inference time and 
the overall end-to-end fusion delay, in addition to comparing the number of processed frames. 
For accuracy, we employ detection completeness to assess the performance of detecting 
all cases against the offline ground truth. Lastly, we examine the cost-effectiveness 
across all cases which is defined as the average fusion latency divided by fusion accuracy;
This is the fusion ratio times the average detection completeness of multiple perception DNN models. 
\begin{align}
\small
\text{Cost-effectiveness} = \frac{\text{average latency}}{\text{fusion percent} 
\times \text{average accuracy}}.
\label{eq:cost_effectiveness}
\end{align}
This cost-effectiveness ratio highlights the \name's efficiency in enhancing 
both latency and accuracy.

\subsection{Multi-Tenant DNN Inference Fusion}
\label{subsec:multi-dnn-fusion}

Since \name\ is designed to make the perception pipeline 
predictable, the multi-tenant DNN inference fusion 
performance becomes our primary goal. Below we discuss 
the fusion performance in terms of the number of fusion 
frames, the fusion delay, and the worst-case fusion delay.

\vspace{0.25em}
\noindent\textbf{Fusion Frames.} The number of fusion 
messages reflect the system's ability to process high 
FPS sensor data. We use two datasets: the BDD100K dataset 
containing 2,629 images and the nuScenes mini v1.0 dataset 
with 1,938 images as the workload of the perception pipeline 
and collect the number of fusion and processed frames 
for all the testing cases. 
The results are presented in 
Tables~\ref{tab:one-A10-num-frames-bdd100k}
and \ref{tab:one-A10-num-frames-nuscenes}, showing that 
the baseline's fusion frames over all the frames (2,629) 
is just 8.6\%. The introduction of the frame dispatcher and 
critical frame selector does not make \name\ process more 
frames than the baseline, because \name\ is designed to 
trade the number of processed frames for better fusion 
performance. The comparison of different test cases shows that 
introducing the detection predictor (DP) greatly increases 
the number of fusion messages because the DP 
publishes prediction detection messages to the fusion
when the detection process is not updated. 
%CCCCC What do you mean by "publishes prediction detection messages"?
%% It means publish the detection prediction results.
This decoupling allows the prediction to run in parallel 
%CCC What does "This decoupling" point to?
%% It means decouping of detection process and fusion process by adding a prediction process.
with the detection, making more messages fused. 
Table~\ref{tab:one-A10-num-frames-bdd100k} shows \name\
to achieve 62.5\% fusion frames compared to the baseline's 
8.6\%,making nearly a 7.3$\times$ improvement. 
Table~\ref{tab:one-A10-num-frames-nuscenes} presents the result on 
the nuScenes dataset, showing \name's achievement of 48\% fusion 
frames compared to the baseline's 20.8\%, which is a 2.3$\times$ 
improvement. The main difference between BDD100K and nuScenes 
datasets is the FPS. At a lower FPS (12Hz), the similarities 
between consecutive frames are lower for the nuScenes dataset. 
However, \name\ still makes a large improvement in the number of 
fusion frames, indicating the effectiveness of the 
detection predictor.

\begin{table}[!htp]
\centering
\caption{The number of processed frames on BDD100K.}
\label{tab:one-A10-num-frames-bdd100k}
\resizebox{\columnwidth}{!}{%
\begin{tabular}{@{}cccccc@{}}
\toprule
\textbf{Frames} & \textbf{Faster R-CNN} & \textbf{DNLNet} & \textbf{Deeplabv3+} & \textbf{Fusion} & \textbf{Percent (\%)} \\ \midrule
Baseline               & 335                  & 286            & 227                & 227            & 8.6                     \\ \midrule
FD                     & 332                  & 282            & 227                & 212            & 8.1                     \\ 
FD+FG                  & 444                  & 491            & 303                & 264            & 10.0                     \\ 
FD+DP                  & 367                  & 322            & 257                & \textbf{1925}  & \textbf{73.2}           \\ \midrule
\name                 & \textbf{459}         & \textbf{536}   & \textbf{322}       & 1642           & 62.5                    \\ \bottomrule
\end{tabular}%
}
\end{table}

\begin{table}[!htp]
\centering
\caption{The number of processed frames on nuScenes.}
\label{tab:one-A10-num-frames-nuscenes}
\resizebox{\columnwidth}{!}{%
\begin{tabular}{@{}cccccc@{}}
\toprule
\textbf{Frames} & \textbf{Faster R-CNN} & \textbf{DNLNet} & \textbf{Deeplabv3+} & \textbf{Fusion} & \textbf{Percent (\%)} \\ \midrule
Baseline               & 612                  & 517            & 410                & 403            & 20.8                     \\ \midrule
FD                     & 609                  & 514            & 410                & 376            & 19.4                     \\ 
FD+FG                  & 679                  & 518            & 446                & 412            & 21.3                     \\ 
FD+DP                  & 604                  & 761            & 412                & \textbf{964}   & \textbf{49.7}            \\ \midrule
\name                 & \textbf{661}         & \textbf{824}   & \textbf{445}       & 930            & 48.0                     \\ \bottomrule
\end{tabular}%
}
\end{table}

% \begin{figure}[!htp]
% 	\centering
% 	\includegraphics[width=.75\columnwidth]{figures/frames-fusion.pdf}
%         % \vspace{-1.5em}
% 	\caption{The number of fused frames for all cases.}
% 	\label{fig:frames-fusion}
% \end{figure}

\begin{figure*}[!htp]
	\centering
	\includegraphics[width=\textwidth]{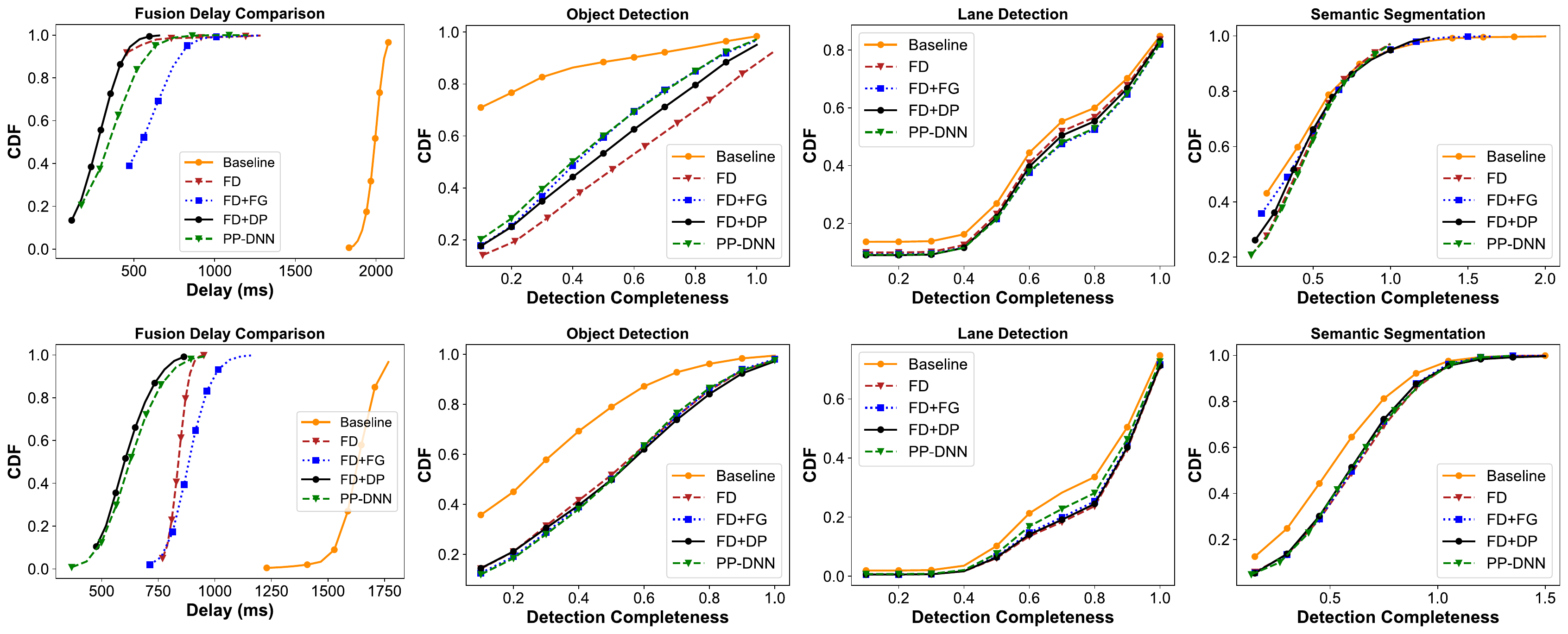}
        \vspace{-2em}
	\caption{The CDF of fusion delay and the detection 
 completeness on BDD100K (upper figures) and nuScenes 
 datasets (lower figures).}
	\label{fig:cdf-cases}
\end{figure*}

\vspace{0.25em}
\noindent\textbf{Fusion Delay.} Another essential metric 
for the AV's safety is the end-to-end delay for the 
the fusion process, defined as the time from capturing the 
image to finish the fusion. 
We collect the fusion delay, inference delay, and image 
sequence number difference for all test cases. 
The results of the fusion delay are shown in 
Fig.~\ref{fig:cdf-cases}. 
% See the Appendix for the CDF for all the tasks in 
% five testing cases.
From Fig.~\ref{fig:cdf-cases}, we can observe that 
\name\ makes a significant improvement in the fusion 
delay over the baseline. 
Table~\ref{tab:one-A10-average-delay-bdd100k} shows 
the average delay for all test cases on the BDD100K dataset. 
\name\ speeds up the fusion by 5.7$\times$ (from 1988.8ms to 349.1ms). 
Table~\ref{tab:one-A10-average-delay-nuscenes} shows the result on 
the nuScenes dataset, showing \name's speedup of the fusion 
by 2.6$\times$ (from 1627.6ms to 632ms). 
By comparing the baseline with FD, we can find that the 
frame dispatcher helps reduce the fusion delay 
by coordinating multi-tenant DNN inference. Besides, from
the comparison of FD with FD+FG, we can observe that introducing 
the critical frame ROI generator reduces the latency.

Table~\ref{tab:fusion-delay-variations} shows the fusion-delay 
variations on both datasets. \name\ significantly reduces the 
variations, primarily thanks to the frame dispatcher, which skips 
out-of-date frames, thereby reducing the maximum delay. 
The detection predictor also contributes by predicting results, 
which reduces both the minimum and maximum fusion delays. We find 
that \name\ narrows the range to 607 ms on the BDD100K dataset, 
which is approximately 2.7$\times$ narrower than the baseline range 
of 1640 ms. Similarly, for the nuScenes dataset, PP-DNN decreases 
the range to 582 ms, which is about 2.3$\times$ narrower than 
the baseline range of 1315 ms.
% \vspace{0.5em}
% \noindent\textbf{Fusion Delay Variations.}

\begin{table}[!htp]
\caption{Fusion-delay variations.}
\centering
\label{tab:fusion-delay-variations}
\resizebox{\columnwidth}{!}{%
\begin{tabular}{@{}cccc|ccc@{}}
\toprule
\multirow{2}{*}{\textbf{\begin{tabular}[c]{@{}c@{}}Fusion Delay\\ Variations (ms)\end{tabular}}} &
  \multicolumn{3}{c|}{\textbf{BDD100K}} &
  \multicolumn{3}{c}{\textbf{nuScenes}} \\ \cmidrule(l){2-7} 
         & min         & max          & range        & min          & max          & range        \\ \midrule
Baseline & 432         & 2072         & 1640         & 417          & 1732         & 1315         \\
\midrule
FD       & 363         & 797          & \textbf{434} & 403          & 908          & 505          \\
FD+FG    & 380         & 857          & 477          & 475          & 1069         & 594          \\
FD+DP    & \textbf{54} & \textbf{538} & 484          & 412          & \textbf{834} & \textbf{422} \\
\midrule
PP-DNN   & 59          & 666          & 607          & \textbf{301} & 883          & 582          \\ \bottomrule
\end{tabular}%
}
\end{table}

\begin{figure}[!htp]
	\centering
	\includegraphics[width=.7\columnwidth]{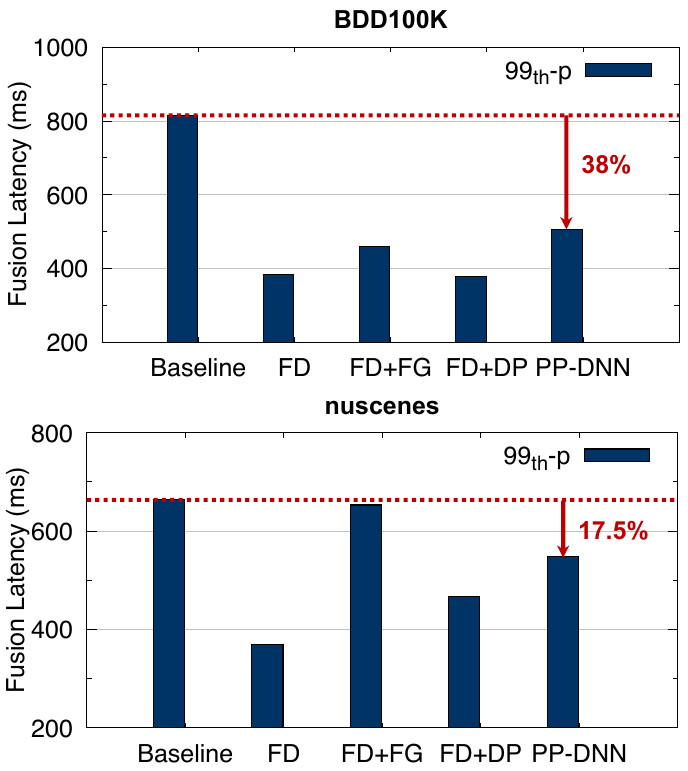}
        % \vspace{-1.5em}
	\caption{The 99-th percentile fusion delay.}
	\label{fig:fusion-delay-99th}
\end{figure}

\vspace{0.5em}
\noindent\textbf{Worst-Case Analysis.} Besides the average 
delay, our comparison of worst-case performance of the baseline and 
\name\ takes into account the 99-th percentile of fusion latency, 
as depicted in Fig.~\ref{fig:fusion-delay-99th}. On the BDD100K 
dataset, we observe a substantial reduction of the worst-case 
fusion delay with \name, from 815ms to 505ms, a decrease of 38\%, 
as opposed to the baseline's delay of $>$800ms. For the nuscenes 
dataset, \name\ reduces the worst-case fusion latency by 17.5\%, 
from 664ms to 548ms. Ablation studies indicate the frame 
dispatcher (FD) helps make these improvements by orchestrating 
multi-tenant DNN inference execution and skipping outdated 
frames. Moreover, the introduction of the frame ROI 
generator (FG) and detection predictor (DP) incurs some 
latency overhead. Nonetheless, \name\ manages to achieve a 
balanced fusion delay, effectively improving worst-case fusion 
latency over the baseline.

% \noindent\textbf{Takeaway 1:} \name\ significantly improves multi-tenant 
% DNN inference. The detection predictor helps increase the number of 
% fused frames by 4.5x. The frame dispatcher helps reduce the fusion 
% delay by 58.9\%. \name\ guarantees the predictability of the 
% perception pipeline.

\begin{table}[]
\centering
\caption{The average delay for tasks on BDD100K.}
\label{tab:one-A10-average-delay-bdd100k}
\resizebox{.8\columnwidth}{!}{%
\begin{tabular}{@{}cccccc@{}}
\toprule
\textbf{Average Delay (ms)} & \textbf{Faster R-CNN} & \textbf{DNLNet} & \textbf{Deeplabv3+} & \textbf{Fusion} & \textbf{Speedup} \\ \midrule
Baseline           & 257.4         & 311.2  & 366.2      & 1988.8  & -      \\ 
\midrule
FD                 & 261.4         & 302.7  & 353.4       & 434.2  & 4.6$\times$   \\
FD+FG              & 261.4         & 247.1  & 356.9       & 570.9  & 3.5$\times$   \\
FD+DP              & \textbf{249.7}	       & 296.6	& \textbf{348.8}	      & \textbf{275.9}  & \textbf{7.2$\times$}   \\ 
\midrule
\name\             & 251         & \textbf{226.6}  & 358.1       & 349.1  & 5.7$\times$   \\ 
\bottomrule
\end{tabular}%
}
\end{table}

% Average Delay (ms)	Faster R-CNN	DNLNet	Deeplabv3+	Fusion	Speedup
% Baseline	257.4	311.2	366.2	1988.8	-
% FD	261.4	302.7	353.4	434.2	4.6
% FD+FG	261.4	247.1	356.9	570.9	3.5
% FD+DP	249.7	296.6	348.8	275.9	7.2
% PP-DNN	251.0	226.6	358.1	349.1	5.7

\begin{table}[!htp]
% \vspace{-0.5em}
\centering
\caption{The average delay for tasks on nuScenes.}
\label{tab:one-A10-average-delay-nuscenes}
\resizebox{\columnwidth}{!}{%
\begin{tabular}{@{}cccccc@{}}
\toprule
\textbf{Delay (ms)} & \textbf{Faster R-CNN} & \textbf{DNLNet} & \textbf{Deeplabv3+} & \textbf{Fusion} & \textbf{Speedup} \\ \midrule
Baseline           & 267.7	& 308.9	& 367.9	& 1627.6  & - \\ 
\midrule
FD                 & 247.8	& 309.5	& 362.2	& 838.5	& 1.9$\times$  \\
FD+FG              & \textbf{238.8}	& 216.4	& 350.6	& 888.3	& 1.8$\times$  \\
FD+DP              & 262.3	& 308.1	& 357.5	& \textbf{610.5}	& \textbf{2.7$\times$}  \\ 
\midrule
\name\             & 250.3	& \textbf{205.2}	& \textbf{342.4}	& 632.0	& 2.6$\times$   \\ 
\bottomrule
\end{tabular}%
}
% \vspace{-0.5em}
\end{table}

% Average Delay (ms)	Faster R-CNN	DNLNet	Deeplabv3+	Fusion	Speedup
% Baseline	267.7	308.9	367.9	1627.6	-
% FD	247.8	309.5	362.2	838.5	1.9
% FD+FG	238.8	216.4	350.6	888.3	1.8
% FD+DP	262.3	308.1	357.5	610.5	2.7
% PP-DNN	250.3	205.2	342.4	632.0	2.6

\subsection{Detection Completeness}
\label{subsec:detection-completeness}

\name\ leverages the ROIs generator to guarantee the detection's 
performance on accuracy. We propose a detection completeness metric 
to evaluate the streaming detection performance when only a group 
of frames is selected for processing. We first discuss the 
effectiveness of the detection completeness metric and then present 
the results of detection completeness for all testing cases.

% \vspace{0.5em}
% \noindent\textbf{Effectiveness in Detection Completeness.} 
% To show the effectiveness in detection completeness on streaming 
% detection, we sampled the same video with different FPS (5, 10, 15) and 
% collected the detection results for each frame. Then, we calculated the 
% IoU for each object on each frame for different FPS. The results for the 
% detection completeness and missed objects compared to the original stream 
% (30 FPS) are shown in Fig.~\ref{fig:fps-detect-cdf}. We can find 
% that the detection completeness reflects the accuracy of the detection 
% of missing objects more clearly than the SSIM and box IoU.

% \begin{figure}[!htp]
% 	\centering
% 	\includegraphics[width=\columnwidth]{figures/insights/box-completeness-fps.pdf}
%         % \vspace{-2em}
% 	\caption{\revise{The CDF of detection completeness ratio 
%  and the number of missed objects when the FPS is 5, 10, and 15.}}
% 	\label{fig:fps-detect-cdf}
% \end{figure}

% \vspace{0.5em}
% \noindent\textbf{Detection completeness of all testing cases.} 
We collected the message sequence numbers for all five testing 
cases and compared them with the offline detection results. 
Also, We collected the detection results for all three tasks when 
running offline as the ground truth. When \name\ is running online, 
we collected detection results and calculated detection 
completeness based on Algorithm~\ref{algorithm1} for each case. 
% For streaming detection, we should also consider the inference time.
% By adding the average inference time, we apply a sequence addition to 
% the ground truth.
%CCCC What's "sequence addition"?
% It's used to add inference time into the evaluation of detection accuracy, adding sequence numbers to get the ground truth
%CCCC "sequence addition" still doesn't make sense
% The addition values to Faster R-CNN, DNLNet, and Deeplabv3+ are 3, 6, 
% and 6, respectively. Because their average inference time is 100ms, 
% 200ms, and 200ms, and the image is published every 33ms. 
The results for the CDF of detection completeness ($d_c$) for all 
testing cases are shown in Fig.~\ref{fig:cdf-cases}. 
For both BDD100K and nuScenes dataset, 
we can observe \name\ improves object detection 
completeness and maintains the same level of detection completeness 
for lane detection and semantic segmentation. 
The improvement mainly comes from our implementation of 
an object tracker for updating the bounding boxes of tracked objects. 
Table~\ref{tab:one-A10-detection-metric-bdd100k} includes the 
average detection completeness for all the testing cases on BDD100K 
dataset. \name\ is found to improve the detection 
completeness by 186\% (from 0.147 to 0.422). This improvement 
is mainly achieved by FD which skips out-of-date frames.
Table~\ref{tab:one-A10-detection-metric-nuscenes} includes the 
average detection completeness for all the testing cases 
on the nuScenes dataset. \name\ is found to improve 
the detection completeness by 75.4\% (from 0.275 to 0.483). 
It also improves the detection completeness for segmentation
by 19.9\% (from 0.488 to 0.585). We can observe that the frame 
ROI generator (FG) enhances the detection completeness using 
its modeling and process of critical areas.
Overall, the detection completeness is enhanced thanks to the frame 
dispatcher (FD) which skips out-of-date frames and the frame 
ROI generator (FG) which captures critical areas for
detection accuracy.

% \usepackage{booktabs}
% \usepackage{graphicx}
% \begin{table}[]
% \vspace{-0.5em}
% \centering
% \caption{The average detection completeness under BDD100K dataset.}
% \label{tab:detection-metric-bdd100k}
% \resizebox{.7\columnwidth}{!}{%
% \begin{tabular}{@{}cccc@{}}
% \toprule
% \textbf{\begin{tabular}[c]{@{}c@{}}Average Detection \\ Completeness\end{tabular}} & \textbf{Faster R-CNN} & \textbf{DNLNet} & \textbf{Deeplabv3+} \\ \midrule
% Baseline    & 0.261 / -          & 0.759 / -          & 0.418 / - \\
% \midrule
% FD          & 0.394 / +51.0\%         & \textbf{0.832 / +9.6\%}          & 0.421 / +0.7\% \\
% FD+FG       & 0.391          & 0.795          & 0.412 \\
% FD+DP       & \textbf{0.396}          & 0.829          & \textbf{0.426} \\
% \midrule
% \nameS & 0.388 / +48.7\% & 0.793 / +4.5\% & 0.418 / +0\% \\ \bottomrule
% \end{tabular}%
% }
% \vspace{-0.5em}
% \end{table}

\begin{table}[!htp]
\centering
\caption{The detection completeness on BDD100K.}
\label{tab:one-A10-detection-metric-bdd100k}
\resizebox{\columnwidth}{!}{%
\begin{tabular}{@{}ccc|cc|cc@{}}
\toprule
\multirow{2}{*}{\begin{tabular}[c]{@{}c@{}}\textbf{Average Detection} \\ \textbf{Completeness}\end{tabular}} &
  \multicolumn{2}{c|}{\textbf{Faster R-CNN}} &
  \multicolumn{2}{c|}{\textbf{DNLNet}} &
  \multicolumn{2}{c}{\textbf{Deeplabv3+}} \\ \cmidrule(l){2-7} 
         & \textbf{$d_c$}    & \textbf{\% Change} & \textbf{$d_c$} & \textbf{\% Change} & \textbf{$d_c$} & \textbf{\% Change} \\ \midrule
Baseline & 0.147          & -                  & 0.643       & -                  & 0.339       & -                  \\ \midrule
FD       & \textbf{0.558}          & \textbf{+279\%}            & 0.679       & +5.6\%             & 0.393       & +15.9\%             \\
FD+FG    & 0.437          & +196\%            & \textbf{0.705}       & \textbf{+9.5\%}             & 0.380       & +12.1\%             \\
FD+DP    & 0.477 & +224\%   & 0.691       & +7.4\%             & 0.369       & +9.1\%             \\ \midrule
\name\   & 0.422          & +186\%            & 0.703       & +9.3\%             & \textbf{0.399}       & \textbf{17.7\%}                \\ \bottomrule
\end{tabular}
}
\end{table}

\begin{table}[!htp]
\centering
\caption{The detection completeness on nuScenes.}
\label{tab:one-A10-detection-metric-nuscenes}
\resizebox{\columnwidth}{!}{%
\begin{tabular}{@{}crc|rc|rc@{}}
\toprule
\multirow{2}{*}{\textbf{\begin{tabular}[c]{@{}c@{}}Detection \\ Completeness\end{tabular}}} &
  \multicolumn{2}{c|}{\textbf{Faster R-CNN}} &
  \multicolumn{2}{c|}{\textbf{DNLNet}} &
  \multicolumn{2}{c}{\textbf{Deeplabv3+}} \\ \cmidrule(l){2-7} 
 &
  \multicolumn{1}{c}{\textbf{$d_c$}} &
  \textbf{\% Change} &
  \multicolumn{1}{c}{\textbf{$d_c$}} &
  \textbf{\% Change} &
  \multicolumn{1}{c}{\textbf{$d_c$}} &
  \textbf{\% Change} \\ \midrule
Baseline & 0.275          & -                & 0.842          & -               & 0.488          & -                \\ \midrule
FD       & 0.469           & +70.2\%          & \textbf{0.902}          & \textbf{+7.1\%}          & \textbf{0.593}          & \textbf{+21.4\%}          \\
FD+FG    & \textbf{0.480} & \textbf{+74.2\%} & 0.893          & +6.0\%          & 0.583 & +19.4\% \\
FD+DP    & 0.479          & +73.9\%          & 0.898 & +6.6\% & 0.582          & +19.2\%          \\ \midrule
\name\   & 0.483          & +75.4\%          & 0.877          & +4.2\%          & 0.585          & +19.9\%          \\ \bottomrule
\end{tabular}%
}
\end{table}

% \noindent\textbf{Takeaway 2:} The detection completeness effectively 
% evaluates the streaming detection performance. \nameS improves the
% completeness of object detection by 15.9\%. It guarantees the 
% functional aspect of predictability for the perception pipeline.

\subsection{Cost-Effectiveness}
\label{subsec:cost-effectiveness}

\begin{figure}[!htp]
	\centering
	\includegraphics[width=\columnwidth]{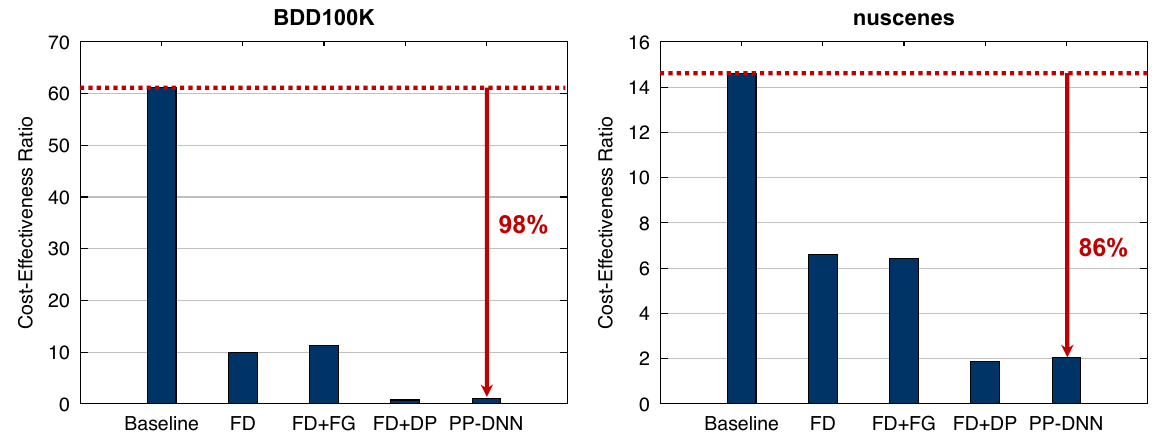}
        % \vspace{-1.5em}
	\caption{The 50-th percentile cost-effectiveness.}
	\label{fig:fusion-cost-effect}
\end{figure}

Fig.~\ref{fig:fusion-cost-effect} shows how effectively 
\name\ enhances both latency and accuracy. 
In particular, this figure details \name's cost-effectiveness 
using various configurations across the BDD100K and nuScenes 
datasets, involving the Frame ROI Generator (FG), 
Frame Dispatcher (FD), and Detection Predictor (DP).

For the BDD100K dataset, by integrating FG, FD, and DP, \name\ 
boosts its cost-effectiveness, showcasing a 98\% (from 61.2 to 
1.1) improvement over the baseline. This underscores the combined 
strategy's role in striking a balance between latency and accuracy. 
For the nuScenes dataset, \name\ demonstrates a notable 
performance advantage with 
an 86\% (from 14.6 to 2.0) enhancement of cost-effectiveness. 
% These figures are accentuated by red dotted baselines 
% and the downward arrows, highlighting the significant improvements with \name's holistic approach.

FD contributes to this increase in cost-effectiveness by skipping 
outdated frames and facilitating coordinated multi-tenant DNN 
inference. The integration of FG also improves this metric by 
selecting critical frames, thereby enhancing detection accuracy 
without processing every frame. Additionally, DP introduces 
predictions for non-critical frames, improving the system's 
efficiency in handling fusion latency. These results suggest that 
while FG, FD, and DP each play a significant role in enhancing 
performance, their collective implementation within \name\ leads to 
the most substantial improvements in cost-effectiveness on 
both datasets.

% \subsection{Memory-Consumption Overheads}

% To evaluate the memory-consumption overhead, we collect 
% both CPU and GPU memory usage for the baseline and \name\. 
% From Table~\ref{tab:memory-consumption}, \name\ is found to 
% use 20\% more CPU memory and 39\% more GPU memory than 
% the baseline. From the breakdown of GPU memory in 
% Fig.~\ref{fig:gpu-memory-breakdown}, we can find that the 
% tracker in \name\ incurs the memory overhead.

% \begin{table}[!htp]
% \centering
% \caption{Comparison of memory consumption}
% \label{tab:memory-consumption}
% \resizebox{.55\columnwidth}{!}{%
% \begin{tabular}{@{}ccc@{}}
% \toprule
% \textbf{Memory (GB)} & \textbf{CPU memory}     & \textbf{GPU memory}   \\ \midrule
% Baseline    & 15.7           & 12.2         \\
% \name\      & \textbf{18.9 (+20.3\%)} & \textbf{17 (+39.3\%)} \\ \bottomrule
% \end{tabular}%
% }
% \end{table}

% \begin{figure}[!htp]
% 	\centering
% 	\includegraphics[width=.8\columnwidth]{figures/gpu-memory-breakdown.pdf}
%         % \vspace{-1.5em}
% 	\caption{The GPU memory breakdown.}
% 	\label{fig:gpu-memory-breakdown}
% \end{figure}

% \section{Discussion}
% \input{contents/Discussion}

\section{Related Work}
\label{sec:related-work}

% \vspace{0.5em}
\noindent\textbf{Real-time DNN Inference.} Extensive work has 
been done on real-time DNN inference. One direction of this 
work was to make the DNN models lightweight. 
Han \textit{et al.}~\cite{han2015learning} proposed the pruning 
of redundant connections to reduce computation demand. 
Reducing the precision of operations and operands is
another direction for the runtime optimization of DNN 
inference~\cite{sze2017efficient,cai2017deep}. 
However, this usually incurs a non-negligible loss of 
accuracy, which is unacceptable for AVs. 
The third direction of work is runtime system acceleration 
and scheduling of the execution pipeline. 
DeepCache~\cite{xu2018deepcache} leverages the temporal 
locality in streaming video to accelerate the vision tasks 
on mobile devices. ALERT~\cite{wan2020alert} addresses this 
with the anytime DNN system, which has multiple outputs 
at different times. Liu \textit{et al.}~\cite{liu2022self} proposed a self-cueing attention mechanism that processes critical regions in real time. However, none of them can support 
real-time multi-tenant DNN inference for AV perception. 
Prophet~\cite{liuprophet} provides temporal predictability
for multi-tenant DNNs without considering functional predictability. 
\name{} is the first to address both aspects of predictability 
with multi-tenant DNN inference for AV perception.

\vspace{0.25em}
\noindent\textbf{Selection of Video Keyframes.} Video keyframe 
detection is commonly used in video detection. 
Xiong \textit{et al.} \cite{xiong2019less} proposed a 
learning-based approach to highlight detection with a preference 
for short videos. Yan \textit{et al.} \cite{yan2018deep} 
proposed a deep two-stream ConvNet for keyframe detection 
in human action videos. Other approaches take multiple input 
sources for keyframe detection: \cite{narasimhan2021clip} 
takes language as additional input while
\cite{badamdorj2021joint} takes audio as additional input. 
However, all these target scenarios where the video is provided 
and the detection is done offline~\cite{badamdorj2022contrastive}. 
In contrast, detecting keyframes for AV perception is more 
challenging since whether a frame is a keyframe or not should 
be decided in real-time, and only information about the 
past frames is available. 

\section{Conclusion}
\label{sec:conclusion}

Predictability of perception latency and accuracy is essential for AV 
safety. The use of multi-tenant DNNs in the AV perception 
pipeline imposes high computational demands, making it even more 
challenging to guarantee both temporally and functionally 
predictability. We have addressed the predictability of AV perception 
with a novel approach by reducing the number of image frames to be 
processed without losing accuracy by dynamically adjusting critical 
frames and ROIs. We have proposed \name\ to trade the number of 
processed frames for better predictability and detection with 
temporal locality. Our in-depth evaluation of the BDD100K dataset 
and the nuScenes dataset has shown \name\ to improve the 
number of fused frames by 7.3$\times$, reduce the fusion delay by 
more than 2.6$\times$ and fusion-delay variations by more than 
2.3$\times$, improve detection completeness by 75.4\%, and achieve 
up to 98\% cost-effectiveness improvement over the baseline with acceptable memory overhead.

% \newpage
\bibliographystyle{IEEEtran}
\bibliography{main}

% \clearpage
% \section*{Appendix}
% \input{contents/9_Appendix}

% \section{Response}
% \input{contents/10_Response}

\end{document}